\documentclass[10pt,twocolumn,letterpaper]{article}

\usepackage{iccv}
\usepackage{times}
\usepackage{epsfig}
\usepackage{graphicx}
\usepackage{amsmath}
\usepackage{amssymb}

\usepackage{graphicx}
\usepackage{amsmath}
\usepackage{amssymb}
\usepackage{booktabs}
\usepackage{color}
\usepackage{caption}
\usepackage{epsfig}
\usepackage{graphicx}
\usepackage{amsmath}
\usepackage{verbatim}
\usepackage{amssymb}
\usepackage{bm}
\usepackage{multirow}
\usepackage{booktabs}
\usepackage{tabularx}
\usepackage{url}
\usepackage{graphicx}
\usepackage{mathrsfs}
\usepackage{amsmath}
\usepackage{multirow}
\usepackage{rotfloat}
\usepackage{algorithmic}
\usepackage{algorithm}
\usepackage{graphicx}
\usepackage[table,xcdraw]{xcolor}
\usepackage[normalem]{ulem}
\useunder{\uline}{\ul}{}

\usepackage{bbding}
\usepackage{pifont}
\usepackage{wasysym}
\usepackage[table,xcdraw]{xcolor}

\usepackage[breaklinks=true,bookmarks=false]{hyperref}

\iccvfinalcopy 

\def\iccvPaperID{****} 

\ificcvfinal\fi

\begin{document}

\title{Multi-interactive Feature Learning and a Full-time Multi-modality Benchmark for Image Fusion and Segmentation}

\author{Jinyuan Liu$^{\dag}$,
	 Zhu Liu$^{\ddag}$, Guanyao Wu$^\ddag$, Long Ma$^\S$, Risheng Liu$^{\S}$,  Wei Zhong$^{\S}$, Zhongxuan Luo$^{\S}$, Xin  Fan$^{\S}$\thanks{Corresponding author.}\\	
	\normalsize$^\dag$School of Mechanical Engineering, Dalian University of Technology\\
	\normalsize $^\ddag$School of Software Technology, Dalian University of Technology\\
	\normalsize $^\S$International School of Information Science Engineering, Dalian University of Technology\\
{\tt \small atlantis918@hotmail.com, liuzhu@mail.dlut.edu.cn, xin.fan@dlut.edu.cn}
}
\twocolumn[{%
	\renewcommand\twocolumn[1][]{#1}%
	\maketitle
	\begin{center}
		\centering
		\captionsetup{type=figure}
		\includegraphics[width=.99\textwidth]{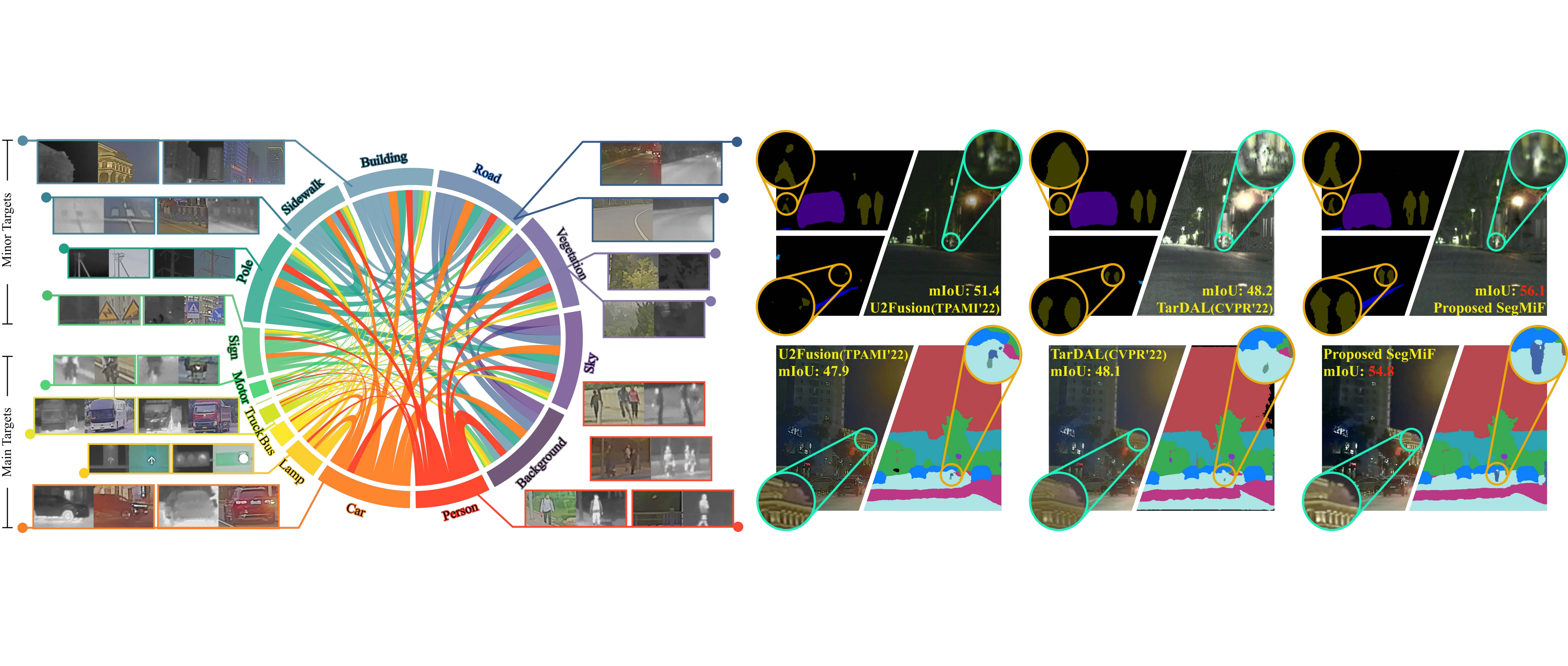}
		\captionof{figure}{The chord diagram on the left shows the association and relative number of various labels in the proposed FMB dataset. Its branches display part of the well registered targets, which are extraordinarily affluent and manifold. The zoomed-in regions on the right show the segmentation comparison in cyan circles and fusion comparison in orange circles. It is obvious that the proposed method is superior to SOTA methods on both visual effects and mIoU.}
		\label{fig:data}
	\end{center}%
}]
\maketitle
\ificcvfinal\thispagestyle{empty}\fi

\begin{abstract}
	\vspace{-0.3cm} 
Multi-modality image fusion and segmentation play a vital role in autonomous driving and robotic operation. Early efforts focus on boosting the performance for only one task, \emph{e.g.,} fusion or segmentation, making it hard to reach~`Best of Both Worlds'. To overcome this issue, in this paper, we propose a \textbf{M}ulti-\textbf{i}nteractive \textbf{F}eature learning architecture for image fusion and \textbf{Seg}mentation, namely SegMiF, and exploit dual-task correlation to promote the performance of both tasks. The SegMiF is of a cascade structure, containing a fusion sub-network and a commonly used segmentation sub-network. By slickly bridging intermediate features between two components, the knowledge learned from the segmentation task can effectively assist the fusion task. Also, the benefited fusion network supports the segmentation one to perform more pretentiously. Besides, a hierarchical interactive attention block is established to ensure fine-grained mapping of all the vital information between two tasks, so that the modality/semantic features can be fully mutual-interactive. In addition, a dynamic weight factor is introduced to automatically adjust the corresponding weights of each task, which can balance the interactive feature correspondence and break through the limitation of laborious tuning. Furthermore, we construct a smart multi-wave binocular imaging system and collect a full-time multi-modality benchmark with 15 annotated pixel-level categories for image fusion and segmentation. Extensive experiments on several public datasets and our benchmark demonstrate that the proposed method outputs visually appealing fused images and perform averagely $7.66\%$ higher segmentation mIoU in the real-world scene than the state-of-the-art approaches.  The source code and benchmark are available at \url{https://github.com/JinyuanLiu-CV/SegMiF}.
\end{abstract}

\section{Introduction}
\label{sec:intro}
Accurate and robust scene parsing\cite{zhao2017pyramid,zhou2017scene} is a fundamental technology for autonomous driving. However, in complex environments, \emph{e.g.,} inclement weather, only using visible sensors may fail to accurately recognize targets.  On the contrary, infrared sensors are free from the aforementioned issues but limited in low spatial resolution. Consequently, fusing the infrared and visible image \cite{zhou2021mffenet,zhou2022edge,jiang2022towards,abs-2211-14461,abs-2303-06840}~has become a mainstream solution for better scene understanding.  

Multi-modality fusion for scene parsing needs to provide:~(i).~\emph{robust visual appealing image}: they require continually generating high-quality images in dynamic scenes. (ii).~\emph{accurate semantic segmentation}: they demand to assign category labels to each pixel.  Towards these goals, jointly solving multi-modality image fusion and segmentation becomes an urgent issue.

Numerous learning-based multi-modality image fusion methods have been fast development~\cite{MFEIF2021,U2Fusion2020,zhao2020didfuse,reconet,UMFusion}. However, most of them concentrate on developing various networks for generating visual-appealing images rather than considering the follow-up high-level vision tasks, posing an obstacle to better scene parsing. Recently, few studies\cite{TarDAL,SeaFusion,ma2022toward,liu2021learning,liu2022twin} have attempted to design multi-task learning-based loss functions by cascading the fusion network and high-level tasks. Unfortunately, seeking unified appropriate features for either task simultaneously is still a tough issue. 


Moreover,~exploring multi-modality fusion and segmentation demands a comprehensive collection of well-alignment image pairs with pixel-level annotated labels. Also, as for one image, the annotated needs to cover a wide range of pixels. Unfortunately, existing multi-modality data collections either focus on image fusion or lack whole image annotated segmentation labels, placing an obstacle to exploring the correlation of the fusion and segmentation.  

This paper proposes a multi-interactive feature learning architecture for the joint problem of multi-modality fusion and segmentation, namely SegMiF. SegMiF is constructed by a fusion network and a segmentation network, in which the intrinsic features of either one interact via a new proposed hierarchical interactive attention ~(HIA). HIA fully integrates semantic-/modality-oriented features by fine-grained mapping. We also derive a dynamic weighting factor and seamless it in the interactive training scheme, to automatically learn the optimal parameters for either task. Figure~\ref{fig:data} demonstrates that our SegMiF assigns the category to each pixel from the visual-friendly fused result more accurately than the state-of-the-arts~(SOTAs). Our contributions can be distilled into four main aspects as follows:
\begin{itemize}
	\item We formulate both image fusion and segmentation in a joint manner, in which the semantic and pixel-based features can mutually interact. To this end, two tasks can achieve the `Best of Both Worlds', generating visual-appealing fused images along with accurate scene parsing.
	\vspace{-0.2cm} 
	\item A hierarchical interactive attention  is introduced to bridge the feature gap between the fusion network and the segmentation one. Establishing the semantic/modality multi-head attention mechanism in HIA simultaneously preserves intrinsic modality features and brings more attention to semantic features.
	\vspace{-0.3cm} 	
	\item An interactive feature training scheme is proposed to overcome the shortcoming of insufficient feature interaction between fusion and segmentation. Seamlessly integrating a dynamic weighting factor allows the exploration of the optimal parameters of each task in an automatic manner.
	
	\vspace{-0.3cm} 
	\item We construct a smart multi-wave binocular imaging system, and introduce a full-time multi-modality benchmark, namely FMB, to promote the research of both image fusion and segmentation. FMB contains 1500 well-registered infrared and visible image pairs with 15 annotated pixel-level categories~(see the left part of Figure~\ref{fig:data}). Also, it covers a wide range of pixel variations and various severe environments, \emph{e.g.,} dense fog, heavy rain, and low-light condition.
	
\end{itemize}
\section{Related Works}

\noindent\textbf{Multi-modality image fusion} 
In recent years, deep learning based multi-modality image fusion approaches achieved significant progress\cite{reconet,UMFusion,zhang2023ingredient,huang2023learning,liu2023holoco}. Early efforts~\cite{MFEIF2021,U2Fusion2020,zhao2020didfuse,liu2023bilevel,liu2021searching} tend to achieve excellent fusion effects by adjusting the network structure or loss functions. However, a minority pay attention to whether the downstream tasks can be well adapted to fusion. Recently, some methods~\cite{wang2023interactively,TarDAL,li2023lrrnet} cascaded fusion and downstream tasks, focusing on improving task performance by achieving oriented fusion. Nevertheless, this kind of native gradient back propagation hinders the fusion network to adapt to subsequent tasks heuristicly from the feature level.

\noindent\textbf{Multi-modality Segmentation}
Recently, two-stream-based  feature fusion models are proposed to perform the segmentation directly.  Most of existing methods mostly develop various simple fusion strategies, such as the weighted average~\cite{guan2019fusion,zhang2020revisiting,liu2023task}, summation~\cite{sun2019rtfnet,zhou2022effective} and concatenation~\cite{ha2017mfnet,shivakumar2020pst900}. Recently, Zhang~\emph{et al.}~\cite{abmdrnet} proposed the multi-scale spatial/channel context modules to fuse features from diverse backbones. Zhou~\emph{et al.}~\cite{zhou2021edge} introduced the densely cascaded contextual inception module to fuse features. Nonetheless, straightforward feature fusion lacks explicit 
fusion principle to preserve the typical modality feature and pay no attention to the pixel-level visual effects.
\begin{figure*}[!htb]
	\centering
	\setlength{\tabcolsep}{1pt} 
	
	\includegraphics[width=0.99\textwidth]{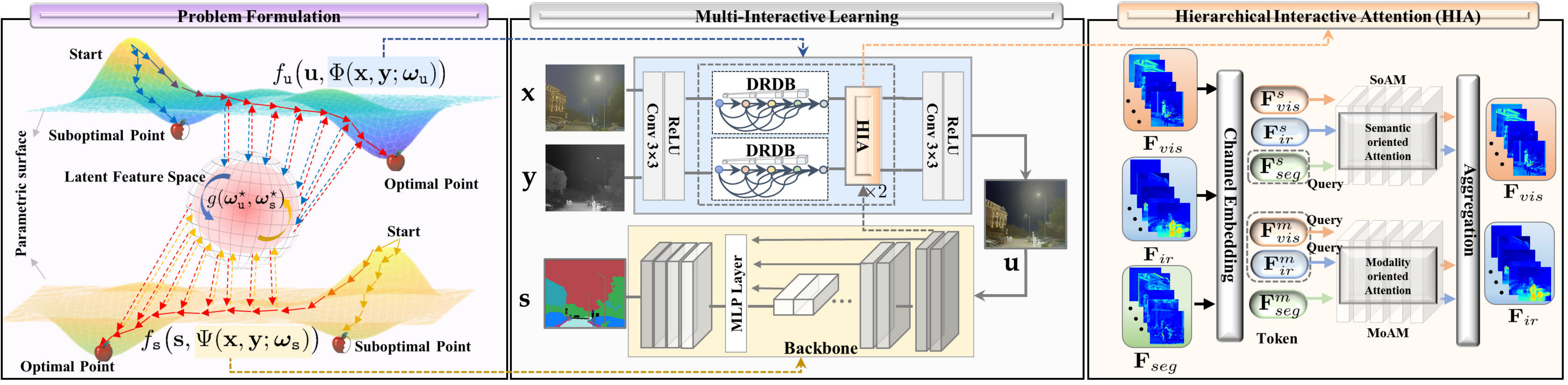}
	\caption{Workflow of our proposed SegMiF. The left part depicts the latent interactive relationship between image fusion and segmentation. The middle part plots the concrete architecture of the SegMiF. The right part  details the components of proposed hierarchical interactive attention. }
	\label{fig:workflow}
\end{figure*}

\section{The Proposed Method}
In this section, we start from the joint formulation of image fusion and segmentation. Then, we bridge the two networks via hierarchical  attention to achieve the interactive aggregation of distinct features. Finally, we introduce a dynamic weighting factor for realizing interactive learning. 
\subsection{Problem formulation}
As for image fusion or segmentation tasks, one of the most commonly used ways is to design a neural network, and fully utilize it to find a set of optimal parameters. For this purpose, we suppose that the visible, infrared,  and fused image are all gray-scale with the size of $m\times n$, denoted as column vectors $\mathbf{x}$, $\mathbf{y}$, and $\mathbf{u}\in\mathbb{R}^{{mn}\times{1}}$,~respectively. The optimization model is formulated as:
\begin{eqnarray}
&\quad\min\limits_{\bm{\omega}_{\mathtt{k}}}f\big(\mathbf{k},\mathcal{N}(\mathbf{x},\mathbf{y};\bm{\omega}_{\mathtt{k}})\big),
\end{eqnarray}
where~$\mathbf{k}$~denotes the output of the task-related network~$\mathcal{N}$ with the learnable parameters~$\bm{\omega}_{\mathtt{k}}$.~$f$($\cdot$)~is a fidelity term.

Previous approaches solely design image fusion or segmentation networks, which only can achieve outstanding results for one task. To generate visual-appealing fused images along with accurate scene segmentation results, we jointly formulate the two tasks into one goal~\cite{liu2021investigating,sun2022detfusion,sun2022drone}, which can be rewritten as: 

\begin{equation}\small
\quad\min\limits_{\bm{\omega}_{\mathtt{u}}, \bm{\omega}_{\mathtt{s}}} f_{\mathtt{u}}\big(\mathbf{u},\Phi(\mathbf{x},\mathbf{y};\bm{\omega}_{\mathtt{u}})\big) + f_{\mathtt{s}}\big(\mathbf{s},\Psi(\mathbf{x},\mathbf{y};\bm{\omega}_{\mathtt{s}})\big) + g(\bm{\omega}^{\star}),
\end{equation}

where $\bm{\omega}^{\star}$ = $[\bm{\omega}_{\mathtt{u}}, \bm{\omega}_{\mathtt{s}}]$. $\mathbf{u}$ and $\mathbf{s}$ denote the fused image and segmentation map, which are produced by the fusion network $\Phi$ and segmentation network~$\Psi$ with the learnable parameters~$\bm{\omega}_{\mathtt{u}}$ and $\bm{\omega}_{\mathtt{s}}$.~$g$~($\cdot$)~is a constrained term to joint optimize the two tasks. In this paper,  we regard the $g$~($\cdot$) as a feature learning constrained manner, and achieve this goal by designing a hierarchical  attention along with the interactive training scheme. The visualized illustration is plotted in the left part of Figure~\ref{fig:workflow}.
\subsection{Feature interaction architecture}
\textbf{Overview of the whole network.} Our proposed SegMiF is designed with cascade principle, composited by image fusion and segmentation sub-network. Details of the whole architecture is shown at Figure~\ref{fig:workflow}.
In specific, we utilize two parallel dilated residual dense blocks (DRDB)~\cite{ZhaoZXLP22,yan2019attention} to extract features from visual and infrared images. SegFormer~\cite{xie2021segformer} is leveraged as the baseline segmentation network to provide semantic parsing.  Two scales of semantic features from the backbone, interpolated with original resolutions are embedded into fusion network. In order to sufficiently realize the semantic information sharing, we propose the hierarchical interactive attention (HIA) to transfer high-level knowledge. 

%
%

\textbf{Hierarchical interactive attention.} After obtaining modality feature $\mathbf{F}_{ir},\mathbf{F}_{vis}$ from fusion network and segmentation feature $\mathbf{F}_{seg}$, we build the HIA to construct the fine-grained mapping of these features and strengthen the mutually beneficial representation. Features including $\mathbf{F}_{ir},\mathbf{F}_{vis}$ and $\mathbf{F}_{seg}$ as inputs, two attention mechanisms  are leveraged to globally exchange intermediate features. Concatenating features from attentions, fresh modality features are generated  based on a residual connection. 

In detail, channel embedding is utilize to decompose modality/semantic features.
We can denote the outputs from the linear embedding as $\{\mathbf{F}_{x}^{s},\mathbf{F}_{x}^{m}\}$ with size $\mathbb{R}^{mn\times C}$, where $x\in\{ir,vis,seg\}$ is under a vector formulation. Instead of utilizing the original self-attention mechanisms directly, we bridge these features with complementary interaction   from different representation subspaces by MoAM and SoAM. 

\begin{figure}[!htb]
	\centering
	\setlength{\tabcolsep}{1pt} 
	
	\includegraphics[width=0.48\textwidth]{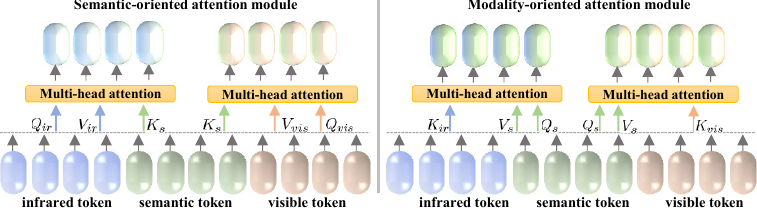}
	
	\caption{Detailed architectures of SoAM and MoAM.}
	\label{fig:SOAMMOAM}
\end{figure}

Both multi-head attention modules are plotted in Figure~\ref{fig:SOAMMOAM}. Semantic-oriented attention module targets to provide more semantic attention for the modality feature. SoAM utilizes the token $\mathbf{F}_{seg}^{s}$ to generate the Query
$Q_{s}$, which represents the inhere semantic information that needs to be enhanced. The global context representation of each can be calculated by as ${K_{ir}}^{T}V_{ir}$ and ${K_{vis}}^{T}V_{vis}$, where the corresponding Key and Value of each head are from the modality tokens $\{\mathbf{F}_{ir}^{m},\mathbf{F}_{vis}^{m}\}$. Denoted the modality context representation as
$\mathbf{G}_{ir}$, $\mathbf{G}_{vis}$, we can calculate the attention as $S_{ir} = Q_{s}\mathbf{G}_{ir}$ and $S_{vis} = Q_{s}\mathbf{G}_{vis}$.
On the other hand, MoAM is with the complementary principle to investigate the significant feature from semantic contexts. Specifically, MoAM is to introduce two modality queries $Q_{ir},Q_{vis}$ to represent the intrinsic modality feature (\textit{e.g.,} targets and details) from $\{\mathbf{F}_{ir}^{m},\mathbf{F}_{vis}^{m}\}$.  Similarly, we can obtain the global context of segmentation $\mathbf{G}_{s}$ by  ${K_{s}}^{T}V_{s}$. The cross attention can be formulated as $M_{vis} = Q_{s}\mathbf{G}_{vis}$ and $M_{vis} = Q_{s}\mathbf{G}_{vis}$. By concatenating the groups $\{S_{vis},M_{vis}\}$ and $\{S_{ir},M_{ir}\}$, we can obtain the comprehensive features with two parallel MLPs with residual connection to aggregate features from SoAM and MoAM.

\subsection{Loss function}
The total loss function is combined of an image fusion loss function~$\mathcal{L}_{\mathtt{f}}$ and segmentation loss~$\mathcal{L}_{\mathtt{s}}$. $\mathcal{L}_{\mathtt{f}}$ consists of three types of losses, \emph{i.e.,} structure loss~$\mathcal{L}_{\mathtt{SSIM}}$, pixel loss~$\mathcal{L}_{\mathtt{MSE}}$ and gradient loss~$\mathcal{L}_{\mathtt{grad}}$. For one fused image, it should preserve overall structures from source images. To this end, the structural similarity index~(SSIM)~\cite{wang2004image,ma2022practical,wu2022breaking,liu2019knowledge} is introduced in function: 
\vspace{-0.2cm} 
\begin{equation}
\mathcal{L}_{\mathtt{SSIM}} = (1-{\mathtt{SSIM}}_{\mathbf{u},\mathbf{x}})/2 + (1 - {\mathtt{SSIM}}_{\mathbf{u},\mathbf{y}})/2, 
\end{equation}
where $\mathcal{L}_{\mathtt{SSIM}}$ denotes structure similarity loss.  To maintains the vital intensity in the fused image, we employ the saliency-based pixel loss, it formulated as :
\begin{equation}
\mathcal{L}_{\mathtt{MSE}} = \lVert \mathbf{u}- \bm{m_1}\mathbf{x}\rVert^2_2 + \lVert \mathbf{u}- \bm{m_2}\mathbf{y}\rVert^2_2,
\end{equation}
where~$\bm{m_1}$~and $\bm{m_2}$ are saliency weight maps calculated by VSM~\cite{ma20171123}. Besides, gradient information of images always characterizes texture details, thus, we used $\mathcal{L}_{\mathtt{grad}}$ to constrain these textual factors to a multi-scale manner:
\begin{equation}
\mathcal{L}_{\mathtt{grad}} = \sum_{k={3,5,7}}\lVert \nabla^k\mathbf{u}- \mathbf{max}(\nabla^k\mathbf{x},\nabla^k\mathbf{y})\rVert^2_2 
\end{equation}
where~$\nabla$ denotes gradient operators that calculate by $\nabla = \mathbf{u}-\mathcal{G}(\mathbf{u})$~with combination of different Gauss~($\mathcal{G}$) kernel size~$k$. 
Totally, we obtained~$\mathcal{L}_{\mathtt{f}} = \mathcal{L}_{\mathtt{SSIM}} + \mathcal{L}_{\mathtt{MSE}} + \eta\mathcal{L}_{\mathtt{grad}}$.

Common to previous works, $\mathcal{L}_{\mathtt{s}}$ is defined as:
\begin{equation}
\mathcal{L}_{\mathtt{s}}(\mathbf{s,s^*}) = -\sum_{class}\mathbf{s^*}\log(\mathbf{s}),
\end{equation} 
where $\mathbf{s^*}$ represents the segmentation label. We adopt the effective semantic segmentation method SegFormer network $\Psi$~\cite{xie2021segformer}. The total loss function is:
\begin{equation}
\mathcal{L}_{\mathtt{total}} = \lambda_1\mathcal{L}_{\mathtt{f}} + \lambda_2\mathcal{L}_{\mathtt{s}},
\end{equation}
where~$\lambda_1$ and $\lambda_2$ are dynamic weighting factor which will be discussed below.

\subsection{Dynamic factor for interactive learning}

\begin{figure*}[!htb]
	\centering
	\setlength{\tabcolsep}{1pt} 
	
	\includegraphics[width=0.95\textwidth,height=0.15\textheight]{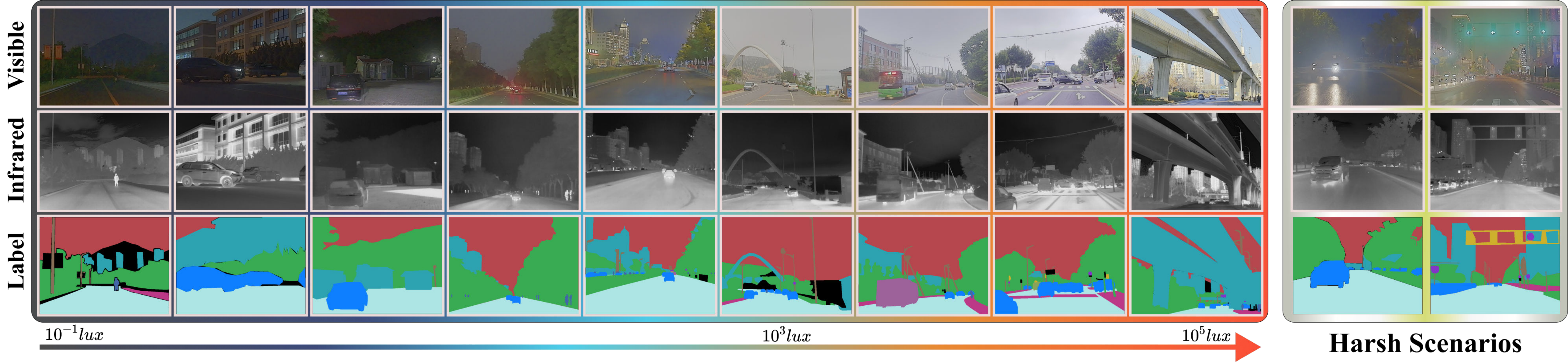}
	
	\caption{Visualization of visible/infrared/segmentation images in the proposed FMB dataset. The dataset contains a wide range of real driving scenes under different lighting conditions, and also includes special scenarios with rain, fog, strong light, and even Tyndall Effect.}
	\label{fig:sample}
	\vspace{-0.5cm}  
\end{figure*}

By exploiting an alternatively recurrent iterations, we can progressively introduce the task-preferred features into the framework optimization. As mentioned above, at the phase of image fusion, we introduce the dynamic weighting factors $\lambda_1$ and $\lambda_2$ to rapidly measure the importance of task-related losses. We observe that the task-specific balance can be derived from the convergent rate.  The intuition is that if the value of losses cannot be further descended, the network may obtain the corresponding optimal weights. If the convergent rate descends fast, we should pay more attention to this task. Learning from the dynamic weight average~\cite{liu2019end}, we further present the factor of task preference $\bm{\eta}$ to emphasize the primary goal. Denoted $r_{i}$ as the convergent rate of $i$-th task with loss $\mathcal{L}_{i}$, we can compute the rate as:
\begin{equation}\label{eq:dyn_r}
r_\mathtt{i} (n-1) = \frac{\mathcal{L}_{i}(n-1)}{\mathcal{L}_{i}(n-2)}.
\end{equation}
Then the procedure of  dynamic weight factor of $i$-th task is formulated as:
\begin{equation}\label{eq:dyn_lambda}
\lambda_\mathtt{i} (n) = \frac{\eta_\mathtt{i}\text{exp}(r_\mathtt{i} (n-1)/T)}{\sum_{k}\text{exp}(r_\mathtt{k} (n-1)/T)},
\end{equation}
where $T$ is a temperature to control the sensitiveness of two tasks. Different from widely sued GDN~\cite{chen2018gradnorm}, this dynamic strategy can  actually avoid the complicated computation of various task gradients. 

Then based on the fusion image generated with semantic feature, we can train the segmentation network end-to-end, using the gradient decending $\bm{\omega}_{s}\leftarrow  \bm{\omega}_{s} - \nabla_{\bm{\omega}_{s} }\mathcal{L}_\mathtt{s}(\mathbf{u};\bm{\omega}_{s})$. The two learning processes are trained alternately until full convergence.
Noting that, this training strategy is actually task-agnostic, we also can introduce other different  high-level vision tasks into the unified consideration, rather than designing for segmentation unilaterally.\footnote{More details of this algorithm  are given in the supplementary material.}

\section{Full-time Multi-modality Benchmark}
\begin{figure}[!htb]
	\centering
	\setlength{\tabcolsep}{1pt}
	\begin{tabular}{c}
		
		\includegraphics[width=0.46\textwidth]{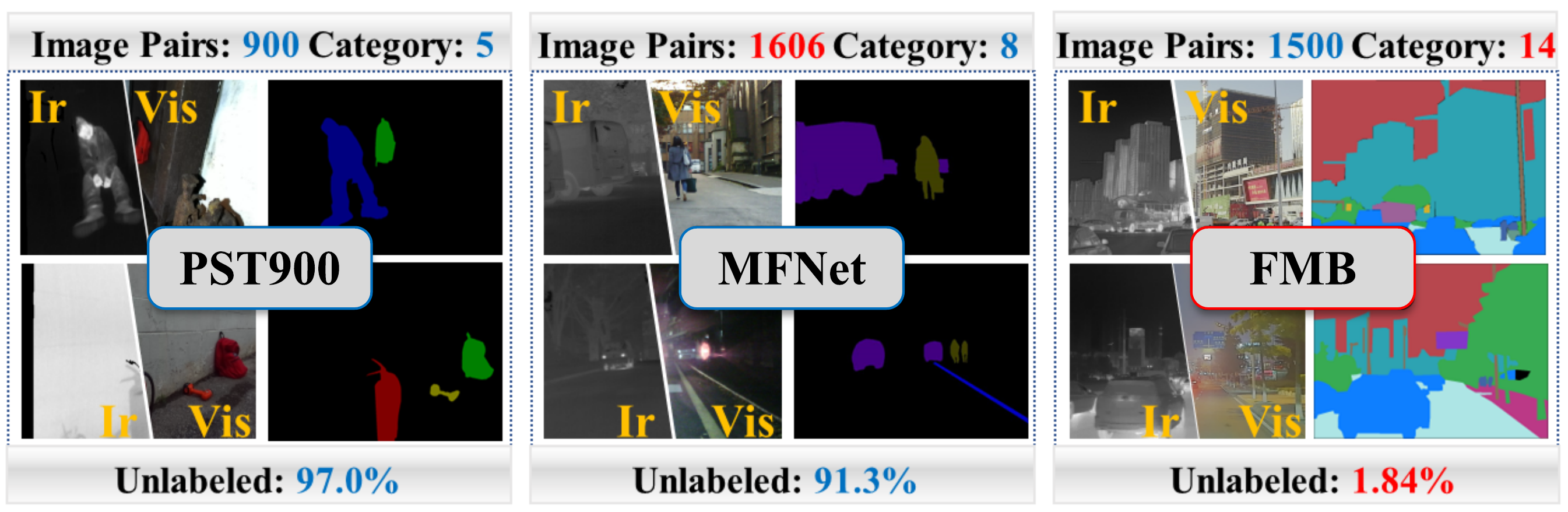}
		\\ 				
	\end{tabular}
	\caption{Comparison of FMB with existing  multi-modality segmentation datasets (\textit{i.e.,} PST900~\cite{shivakumar2020pst900} and MFNet~\cite{ha2017mfnet}).}
	\label{fig:MMSPdatasets}
\end{figure}

Existing two multi-modality segmentation datasets suffer from few label categories, sparse annotation and monotonous scene, as shown in Figure~\ref{fig:MMSPdatasets}. The proposed FMB dataset aims to overcome these difficulties and promote the development of  whole field. A glimpse of FMB is given in Figure~\ref{fig:sample} \footnote{More details  are attached to the supplementary materials.} .

\begin{figure}[!htb]
	\centering
	\setlength{\tabcolsep}{1pt}
	\begin{tabular}{c}
		
		\includegraphics[width=0.46\textwidth]{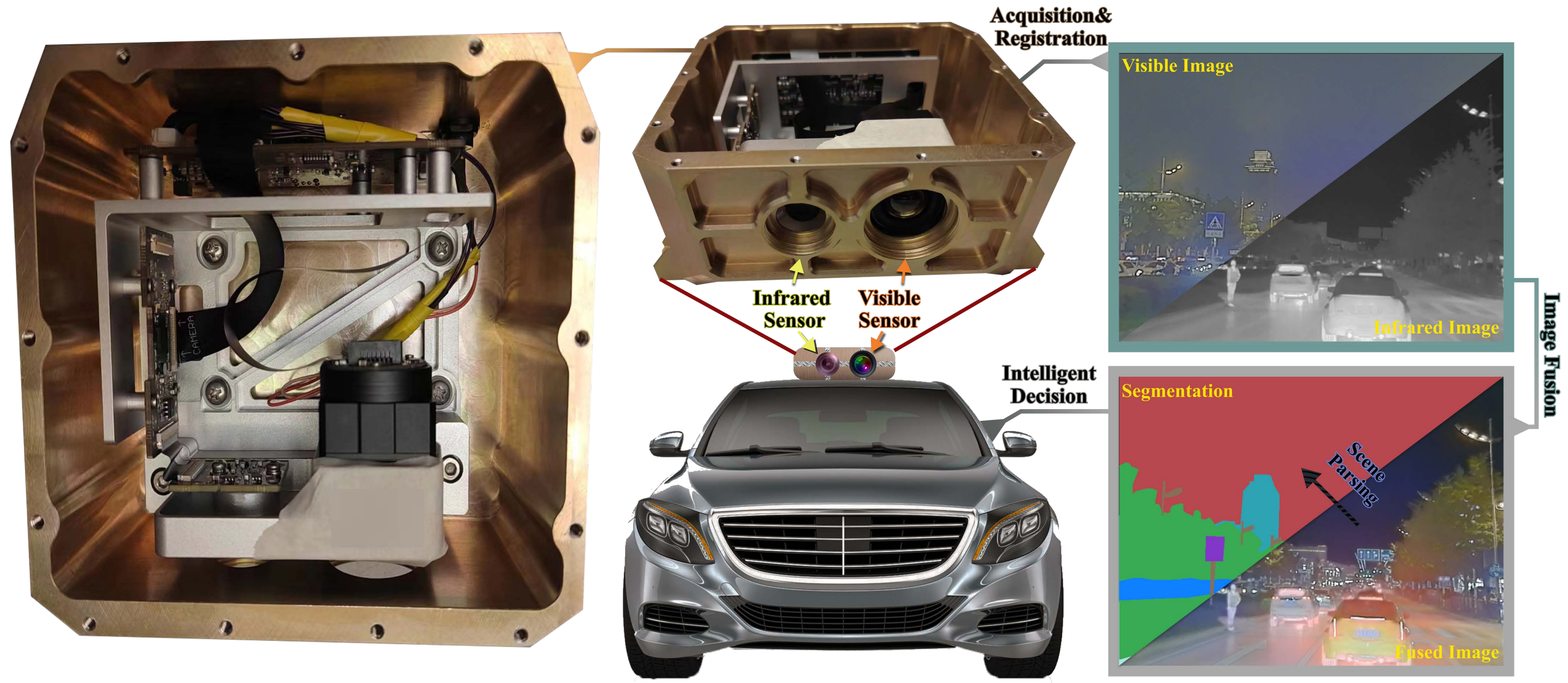}
		\\ 				
	\end{tabular}
	\caption{Illustration of the binocular imaging system.}
	\label{fig:SYSTEM}
\end{figure}

We built a binocular imaging system that can be placed on the car roof, including a visible camera and an infrared sensor with a wavelength range of 8-14$\mu m$(as shown in the Figure~\ref{fig:SYSTEM}). We finally obtained 1500 aligned image pairs with a resolution of 800$\times$600. In details, two sensors are individually calibrated using their respected calibration board to obtain internal parameters, and their relative pose relationship is obtained through joint calibration. We calculate the homography matrix $H$ by employing RANSAC\footnote{Efficient RANSAC for point‐cloud shape detection}. The infrared images are projected onto visible coordinates using $H$ and cropped, ultimately resulting in pixel-level registered image pairs with a size of 800$\times$600. 

The FMB dataset includes rich scenes under different illumination conditions, so that it enables fusion/segmentation model to improve the generalization ability greatly. We labeled 98.16\% of all pixels into 14 different categories including~\emph{Road, Sidewalk, Building, Traffic Light, Traffic Sign, Vegetation, Sky, Person, Car, Truck, Bus, Motorcycle, Bicycle and Pole}, which often appear in real-world automatic driving and semantic understanding tasks. 

\section{Experiments}

\begin{figure*}
	\centering
	\setlength{\tabcolsep}{1pt}
	\begin{tabular}{cccccccc}
		\includegraphics[width=0.12\textwidth,height=0.085\textheight]{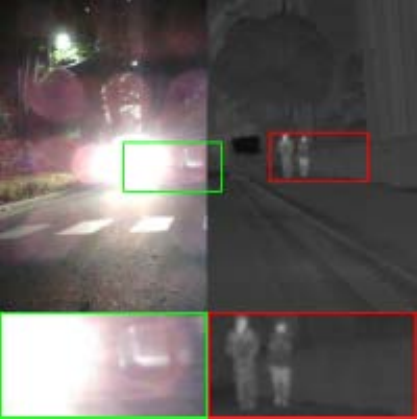}
		&\includegraphics[width=0.12\textwidth,height=0.085\textheight]{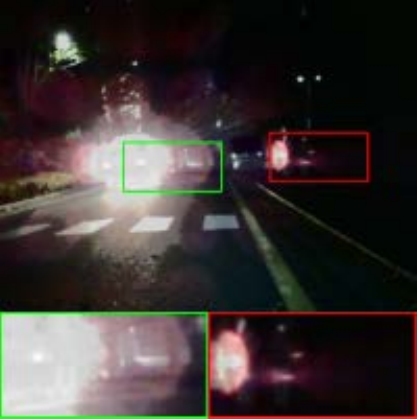}
		&\includegraphics[width=0.12\textwidth,height=0.085\textheight]{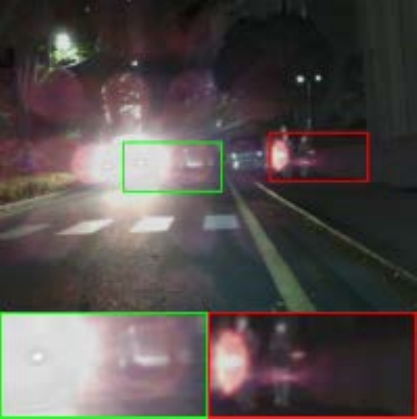}
		&\includegraphics[width=0.12\textwidth,height=0.085\textheight]{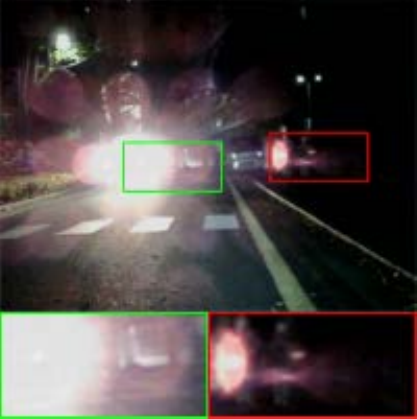}
		&\includegraphics[width=0.12\textwidth,height=0.085\textheight]{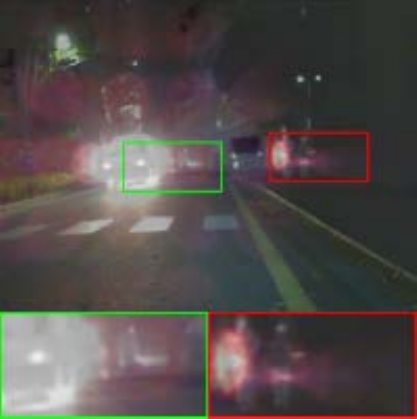}
		&\includegraphics[width=0.12\textwidth,height=0.085\textheight]{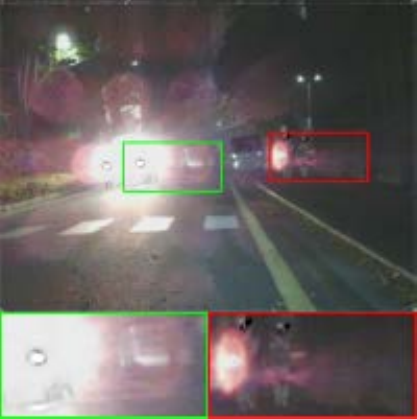}
		&\includegraphics[width=0.12\textwidth,height=0.085\textheight]{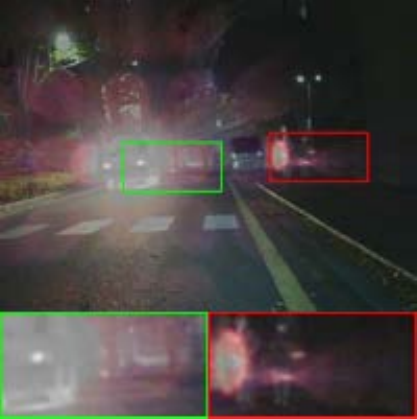}
		&\includegraphics[width=0.12\textwidth,height=0.085\textheight]{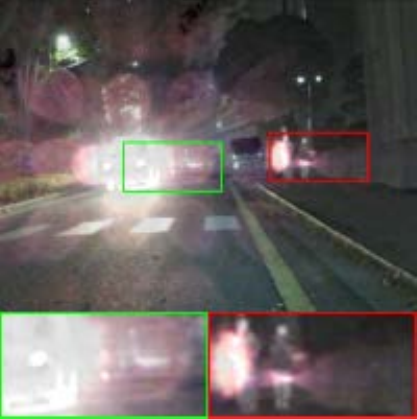}\\
		\includegraphics[width=0.12\textwidth,height=0.085\textheight]{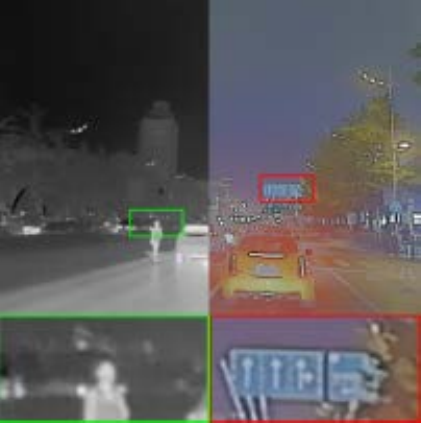}
		&\includegraphics[width=0.12\textwidth,height=0.085\textheight]{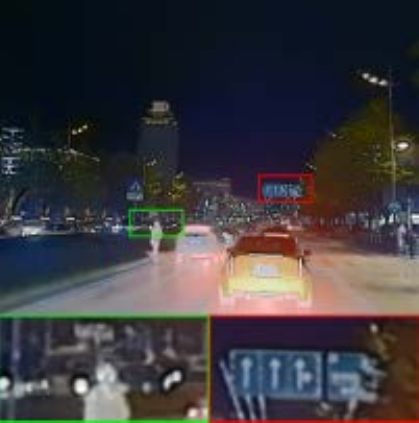}
		&\includegraphics[width=0.12\textwidth,height=0.085\textheight]{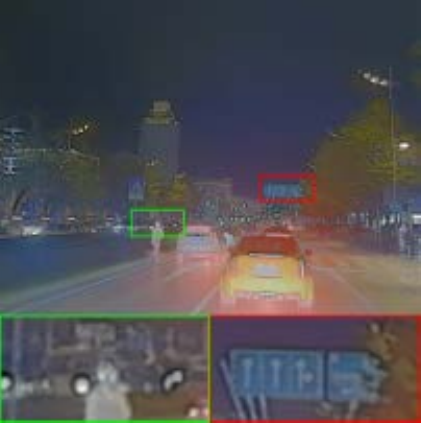}
		&\includegraphics[width=0.12\textwidth,height=0.085\textheight]{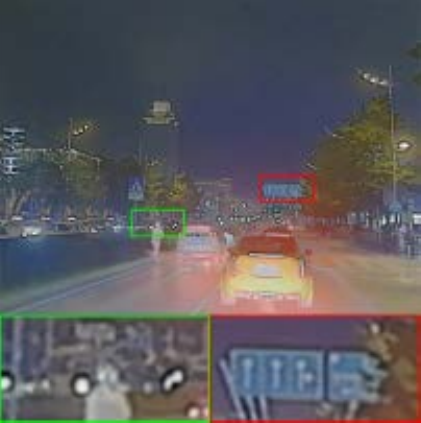}
		&\includegraphics[width=0.12\textwidth,height=0.085\textheight]{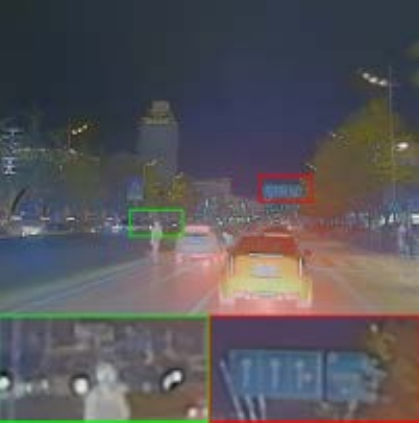}
		&\includegraphics[width=0.12\textwidth,height=0.085\textheight]{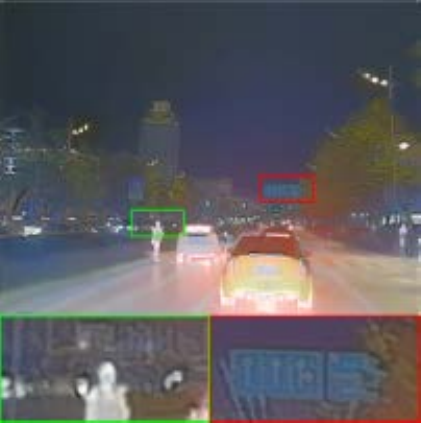}
		&\includegraphics[width=0.12\textwidth,height=0.085\textheight]{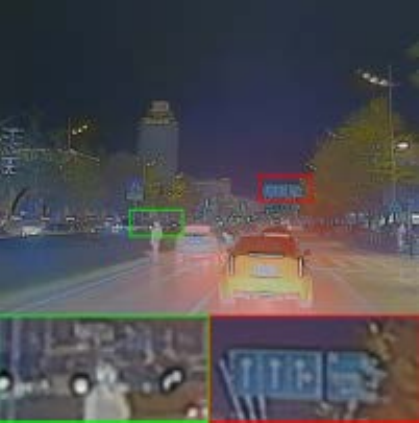}
		&\includegraphics[width=0.12\textwidth,height=0.085\textheight]{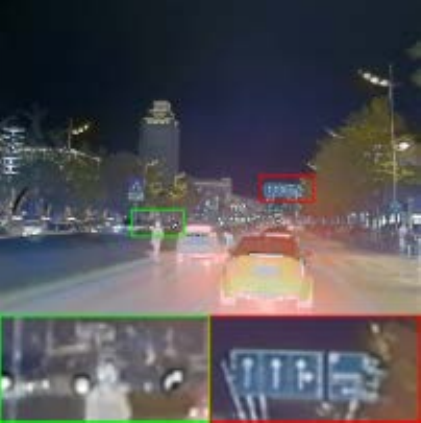}\\
		Source images&DIDFuse&DenseFuse&ReCoNet&UMFusion&TarDAL&U2Fusion&Ours
		\\
	\end{tabular}
	\caption{Visual comparison of different fusion approaches on the MFNet and FMB dataset, respectively.}
	\label{fig:contristive}
\end{figure*}

\begin{figure*}[!htb]
	\centering
	\setlength{\tabcolsep}{1pt} 
	\begin{tabular}{c}		
		\includegraphics[width=0.98\textwidth]{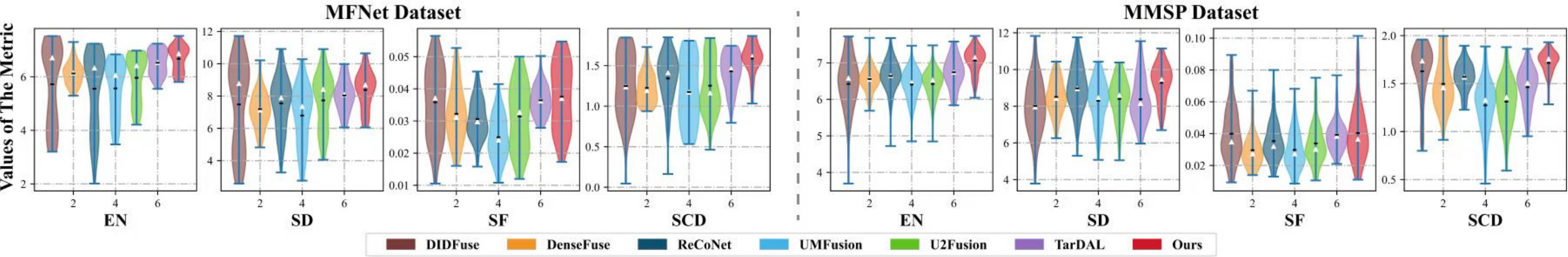}	
	\end{tabular}
	\vspace{-0.3cm}  
	\caption{Quantitative comparisons of image fusion with six SOTA methods on two datasets. Violin plots illustrating the distribution of the four metrics, in which the white triangles and the black lines indicate mean values and medium values.}
	\label{fig:numcpr}
\end{figure*}
\begin{figure*}[!htb]
	\centering
	\setlength{\tabcolsep}{1pt} 
	
	\includegraphics[width=0.98\textwidth,height=0.16\textheight]{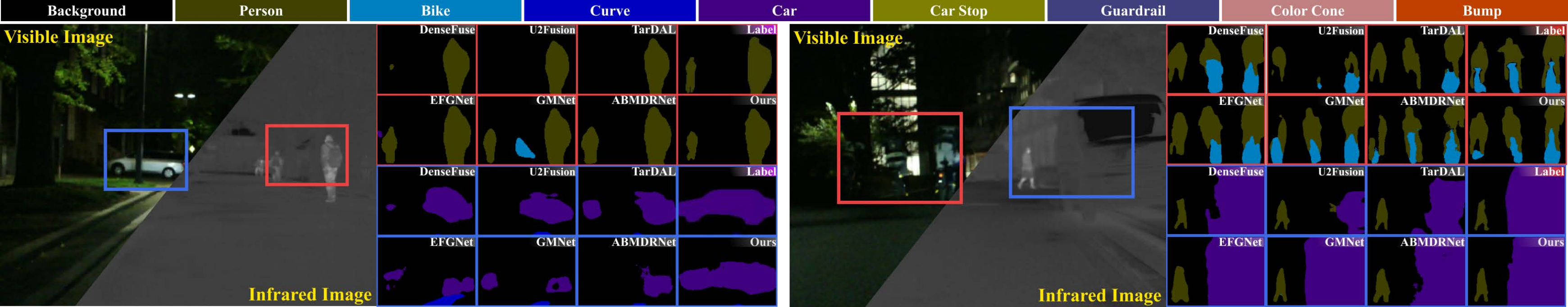}
	
	\caption{Qualitative demonstrations of different approaches on the MFNet dataset.}
	\label{fig:MFresult}
\end{figure*}

Two representative datasets including MFNet and proposed FMB  are utilized for the training and evaluation. The details of these datasets are reported in the above section. Several data augmentation techniques are utilized for the whole training procedure: random resizing with a ratio of 0.5-2.0, random cropping to $360\times360$,  brightness distortion and normalization. The Adam optimizer with poly learning rate adjustment is utilized to optimize both networks. As for the fusion, the initial learning rate is $1e^{-4}$ and decayed to $1e^{-8}$ progressively. As for segmentation, we first utilized $1e^{-6}$ to warm start the training with 3k iterations, then we conducted the training with intilial learning rate $8e^{-5}$.
With batch size of 8, we trained the framework for 8 rounds. For each round, we set 10k iterations for training segmentation and 5k iterations for fusion. Noting that, a similar configuration of segmentation is also utilized for the training of fusion-based methods. All experiments are performed on an NVIDIA Tesla V100 GPU with PyTorch framework.

\subsection{Results of multi-modality image fusion}
We demonstrate our fusion quality based on qualitative and quantitative analyses with six state-of-the-art competitors, including  DIDFuse~\cite{zhao2020didfuse}, DenseFuse~\cite{li2018densefuse}, ReCoNet\cite{reconet}, UMFusion\cite{UMFusion}, TarDAL\cite{TarDAL} and U2Fusion~\cite{U2Fusion2020}. 

\noindent\textbf{Qualitative Comparisons.}
The qualitative results on MFNet and FMB datasets are depicted  in Figure~\ref{fig:contristive}, in which we can clearly observe two remarkable advantages of our method. First, the significant characteristics of infrared images can be effectively highlighted. For instance, as shown on the green rectangle of the first group, DIDFuse, ReCoNet and TarDAL are susceptible to strong illumination. In contrast, our method can remarkably preserve this information from infrared images, \emph{e.g.,} the structure of cars and pedestrians.  Furthermore,  benefiting from the guidance of semantic information, our method can enhance the texture details of the given scene from either dataset. Compared with other competitors in the second row of Figure~\ref{fig:contristive}, our results exhibit a sharper appearance.

\begin{figure*}[!htb]
	\centering
	\setlength{\tabcolsep}{1pt} 
	
	\includegraphics[width=0.98\textwidth,height=0.16\textheight]{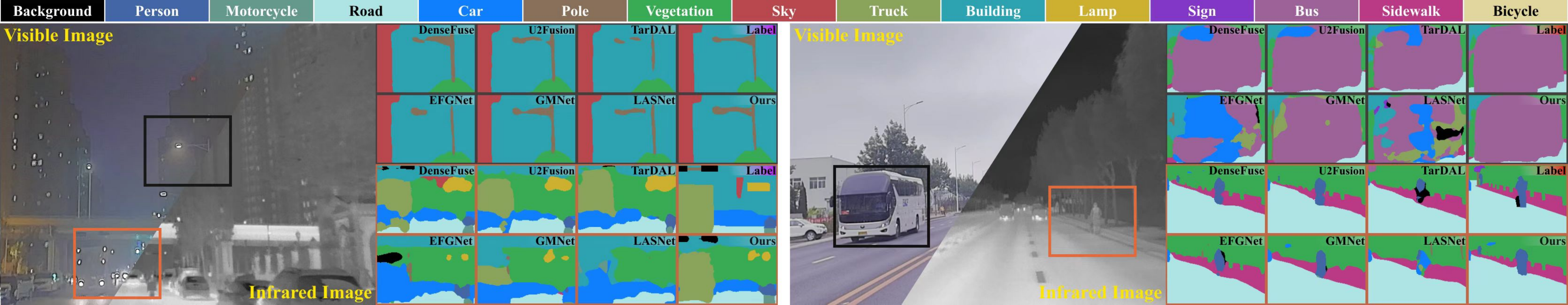}
	
	\caption{Qualitative demonstrations for the different methods in  daytime and nighttime scenarios on FMB benchmark.}
	\label{fig:MSPresult}
\end{figure*}
\begin{table*}[thb]
	\centering
	\footnotesize
	\renewcommand{\arraystretch}{1.1}
	\setlength{\tabcolsep}{1.7mm}{
		\begin{tabular}{|c|cc|cc|cc|cc|cc|cc|cc|cc|cc|}
			\hline
			\multirow{2}{*}{Methods} &			\multicolumn{2}{c}{Unlabel}&  \multicolumn{2}{c}{Car} & \multicolumn{2}{c}{Person} & \multicolumn{2}{c}{Bike}& \multicolumn{2}{c}{Curve}  & \multicolumn{2}{c}{Car Stop}&
			\multicolumn{2}{c}{Cone}& \multicolumn{2}{c|}{Bump}
			&\multirow{2}{*}{mAcc}&\multirow{2}{*}{mIoU} \\ \cline{2-17} 
			&Acc &IoU & Acc &IoU& Acc &IoU& Acc &IoU& Acc &IoU& Acc &IoU& Acc &IoU& Acc &IoU& &\multicolumn{1}{c|}{} \\
			\hline
			Visible  &98.2& 	97.5&94.6	&82.2&81.7 & 54.7&  71.2&	61.0& 54.9&	21.6& 61.9&	16.6&82.1& 	 43.7&77.3&	30.8& 69.7&	45.9
			\\
			Infrared& 98.6&	97.4	&90.3&81.8& 85.3	&67.3& 71.6&	54.8& 51.3&	37.3& 53.8&	24.3& 64.7&	33.4& 70.2	&48.0& 65.6&	49.9\\
			\hline
			LASNet &99.2 &97.4  &94.9  &84.2 & 81.7 &67.1 &\cellcolor{red!15}82.1 &	56.9 &\cellcolor{blue!15}70.7 &41.1 & 56.8 &\cellcolor{red!15}39.6& 58.1 &48.8& 77.2&	40.1  & \cellcolor{red!15}75.4 &54.9 \\
			EGFNet &\cellcolor{red!15}99.3 &97.7 &\cellcolor{blue!15}95.8  &87.6 & 89.0 &\cellcolor{blue!15}69.8 &80.6 &	58.8 & \cellcolor{red!15}71.5 &42.8 & 48.7 &\cellcolor{blue!15}33.8 & 65.3& 48.3 &71.1& 47.1&	72.7 & 54.8\\
			FEANet & 99.1 &97.8 &93.9  &	\cellcolor{red!15}87.8 & 82.7 &71.1 & 76.7 &	\cellcolor{blue!15}61.1 & 65.5 &\cellcolor{blue!15}46.5 &26.6 &22.1 & 66.6 &\cellcolor{red!15}55.3 & 77.3 &	48.9 & 73.2 &\cellcolor{blue!15}55.3 \\
			ABMDR & \cellcolor{red!15}99.3 &\cellcolor{red!15}98.4 &94.3  &	84.8 & \cellcolor{red!15}90.0 &69.6 &75.7 &	60.3 & 64.0 &45.1 & 44.1 &	33.1& 61.7 &	47.4 & 66.2 &	50.0  & 69.5 &	54.8 \\
			\hline
			
			\cellcolor{gray!15}DIDFuse & 98.0&	97.3&95.0&79.1& 79.5&	64.0& 80.2&	58.5	& 44.1&19.9&64.0&	23.3& 77.5&37.8&69.0& 20.4& 67.5&	44.5	 \\
			\cellcolor{gray!15}ReCoNet & 98.1 &97.3 &\cellcolor{blue!15}95.8  &	80.4 & 88.9 &60.0 &65.0 &	55.4 & 47.0 &20.7 & \cellcolor{red!15}69.0 &	25.8& 77.8 &	39.8 & 46.6 &	17.4  & 65.9 &	44.5 \\
			\cellcolor{gray!15}U2Fusion &98.3 &97.7 &95.2  &82.8 & 85.4 &64.8 &77.7 &61.0 & 62.7 &32.3 & 66.7 &20.9 & 75.5& 45.2 &82.3&\cellcolor{blue!15} 50.2&	71.9 & 50.8\\
			\cellcolor{gray!15}TarDAL & 98.3 &97.6 &93.5  &80.7 & 86.2 &67.1 & 76.5 &	60.1 & 53.8 & 34.9 & 55.3 &10.5 & \cellcolor{red!15}88.6 &38.7 & \cellcolor{red!15}90.6 &	45.5 & 71.7 &48.6 \\ 
			
			\cellcolor{yellow!15}\textbf{Ours}  &98.7&\cellcolor{blue!15}98.1 &\cellcolor{red!15}96.3  &	\cellcolor{red!15}87.8 & \cellcolor{blue!15}89.6&\cellcolor{red!15}71.4 & \cellcolor{blue!15}81.2 &	\cellcolor{red!15}63.2 & 63.5 &	\cellcolor{red!15}47.5& \cellcolor{blue!15}66.7 & 31.1 & \cellcolor{blue!15}85.3 &\cellcolor{blue!15}48.9 & \cellcolor{blue!15}84.8 &\cellcolor{red!15}50.3  & \cellcolor{blue!15}74.8 &	\cellcolor{red!15}56.1 \\
			\hline
	\end{tabular} }
	\caption{ Quantitative semantic segmentation results of different methods on the {MFNet} dataset.}~\label{tab:123}
\end{table*}
\begin{table*}[thb]
	\centering
	\footnotesize
	\renewcommand{\arraystretch}{1.1} 
	\setlength{\tabcolsep}{1.7mm}{
		\begin{tabular}{|c|cc|cc|cc|cc|cc|cc|cc|cc|cc|}
			\hline
			\multirow{2}{*}{ Methods} & \multicolumn{2}{c}{ Car} & \multicolumn{2}{c}{  Person} & \multicolumn{2}{c}{Truck}& \multicolumn{2}{c}{T- Lamp}  & \multicolumn{2}{c}{T-Sign}&  \multicolumn{2}{c}{Building}&\multicolumn{2}{c}{Vegetation}& \multicolumn{2}{c|}{Pole}
			& \multirow{2}{*}{mAcc}&\multirow{2}{*}{mIoU} \\ \cline{2-17} 
			& Acc & IoU  & Acc & IoU & Acc & IoU & Acc & IoU & Acc & IoU & Acc & IoU & Acc & IoU& Acc & IoU& &\multicolumn{1}{c|}{} \\
			\hline
			
			Visible  & 84.5	& \cellcolor{blue!15}78.3& 78.1 & 46.6&   73.2&\cellcolor{blue!15}43.4& 82.5& 23.7&  83.6& 64.0& 89.0&  77.8 &88.5 &82.1 & 71.6& 41.8& 73.5& 50.5
			\\
			Infrared	& 75.6& 69.1&\cellcolor{blue!15}89.9	& 63.3&  66.6& 12.6& 63.1& 24.7&  80.5&	 52.9& 87.4& 78.0&83.7 &75.5 &  62.1	& 23.0&  69.6&	 43.9\\
			\hline
			GMNet & \cellcolor{red!15}87.6  & \cellcolor{red!15}79.3 &  77.3 & 60.1 & 49.0 & 	22.2 &  54.1 & 21.6 &  78.5 & \cellcolor{blue!15}69.0&  89.1 & 79.1&88.9 &83.8 &  59.7& 	39.8  &  64.4 & 49.2 \\
			LASNet & 81.3  & 	72.6 & 76.4 & 48.6 & 29.6 & 	14.8 &  20.7 & 2.9 & 79.3 & 59.0& 86.9 & 75.4 &87.6 &81.6 &  56.6& 36.7 & 56.9 & 42.5\\
			
			EGFNet & 83.6  & 77.4 &  79.5& 63.0 & 33.5 & 17.1 &  58.6 & 25.2 &  82.6 & 66.6& 88.5& 77.2 &89.3 &83.5 & 63.8&  41.5& 63.0 &  47.3 \\
			FEANet & 82.3  & 	73.9 &  78.8 & 60.7 &  44.7 & 32.3 &  53.6 & 13.5 & 73.3 & 55.6&  87.6 & \cellcolor{blue!15}79.4&89.0 &81.2 &  66.2 & 36.8 &  64.5& 46.8\\
			
			\hline
			\cellcolor{gray!15}DIDFuse 	& \cellcolor{blue!15}86.3& 77.7&  87.4& 	64.4&  66.3& 	28.8	&  75.9& 29.2& 81.1& 64.4&  87.1& 78.4&89.5 &82.4&\cellcolor{red!15}79.2&  41.8& 73.0& 50.6	 \\
			
			\cellcolor{gray!15}ReCoNet & 83.7& 75.9& 87.7& \cellcolor{red!15}65.8&  34.7& 14.9	&  \cellcolor{blue!15}83.3& 34.7	& \cellcolor{blue!15}85.6& 66.6&  89.0	& 79.2&88.2&81.3	&  73.3&\cellcolor{blue!15}44.9&  71.4& \cellcolor{blue!15}50.9    \\
			
			\cellcolor{gray!15}U2Fusion   & 85.0 &	 76.6&  87.7&  61.9& \cellcolor{red!15}84.6 & 14.4 &  75.1&  28.3 & 81.3&  68.9 &\cellcolor{blue!15}89.5& 78.8&\cellcolor{red!15}92.5 &82.2 &  \cellcolor{blue!15}74.5 & 42.2&  70.1 & 47.9\\ 
			
			\cellcolor{gray!15}TarDAL & 81.8  & 74.2 & \cellcolor{red!15}93.3 & 56.0 &  66.3 & 18.8 &  75.0 &\cellcolor{blue!15}29.6&  81.2 & 66.5	&  88.1 & 79.1 &87.9 &81.7 &  65.9 & 41.9  & \cellcolor{red!15}74.8 & 48.1 \\
			
			\cellcolor{yellow!15}\textbf{Ours}   & 85.3  & \cellcolor{blue!15}78.3 &  78.3 & \cellcolor{blue!15}65.4 & \cellcolor{blue!15}74.4 & 	\cellcolor{red!15}47.3 &  \cellcolor{red!15}86.4 & 	\cellcolor{red!15}43.1 & \cellcolor{red!15}86.1 &  \cellcolor{red!15}74.8& \cellcolor{red!15}90.0 & \cellcolor{red!15}82.0 &\cellcolor{blue!15}91.6 &85.0&  72.5 & \cellcolor{red!15}49.8  &  \cellcolor{blue!15}74.5 & 	\cellcolor{red!15}54.8 \\\hline
	\end{tabular} }
	\caption{ Quantitative semantic segmentation results of different methods on the {FMB} dataset.}~\label{tab:data_set_results_seg}
\end{table*}

\noindent\textbf{Quantitative Comparisons.} We also plot the numerical results with other six fusion competitors on 50 pairs from MFNet and 50 pairs from FMB in Figure~\ref{fig:numcpr}. Four objective metrics are leveraged for the comparison, including entropy (EN)\cite{Roberts2008Assessment}, standard
deviation (SD)\cite{aslantas2015new}, spatial frequency (SF)\cite{Cui2015Detail} and the sum of the correlations of differences (SCD)\cite{zheng2023deep}. Note that our results achieve consistent superiority in terms of these statistical metrics. Specifically, the highest EN and SCD indicate that our method can significantly preserve the largest amount of considerable information transferred from source images. Moreover, the immense average value on SD reflects the high pixel contrast for
visual observations.  Furthermore, higher SF reflects our method has rich texture details and contrasts. 
In summary, our method enhances the texture details for precise observation and stably preserves abundant typical information to support semantic parsing tasks.

\subsection{Results of multi-modality segmentation}
We provide another comprehensive analysis for image segmentation. 
Besides comparing with the newest fusion-based methods, we also conduct the  evaluations with competitive dual-stream methods: GMNet~\cite{zhou2021gmnet}, FEANet~\cite{deng2021feanet}, EGFNet~\cite{zhou2021edge}, ABMDRNet~\cite{abmdrnet} and LASNet~\cite{li2022rgb}.\footnote{We also retrained dual-stream methods on the FMB dataset.}

\noindent\textbf{Qualitative Comparisons.} 
Visualized results of segmentation on MFNet are depicted in Figure~\ref{fig:MFresult}.   We also compare various competitive methods in Figure~\ref{fig:MSPresult} under the newly proposed dataset, which is more challenging with rich categories, complex imaging conditions and complicated scene details.
As discussed above, existing fusion methods cannot highlight the dimness of infrared targets, and the distant pedestrian can not be recognized.
As for dual-stream methods, which utilize the modality feature directly, they are easy to introduce conflicts and weaken the accuracy without a clear feature fusion principle. The results, such as the car occluded by barriers (the first column of Figure~\ref{fig:MFresult})  and the shape of the human (the second column of Figure~\ref{fig:MSPresult}) can not be precisely classified.
It is worth mentioning that interaction feature learning from segmentation can drastically transfer  the complementary  characteristics for image fusion and further improve the segmentation performance.  Thus, our method can continuously classify the objects of diverse scenes with high accuracy.

\noindent\textbf{Quantitative Comparisons.} 
Table~\ref{tab:123} and Table~\ref{tab:data_set_results_seg} reported the qualitative results among different categories of competitors on MFNet and FMB datasets. These results illustrate our method is ahead of other state-of-the-art methods by a large margin on both segmentation datasets. Noting that our numerical results outperform other methods concerning mIOU and rank second in terms of mACC. Compared with the second one, our method improves 7.66\% and 1.45\% of mIOU on FMB and MFNet respectively. More specifically, the classification of Car and Person is important for the current intelligent perception system. The top two results in these two categories indicate the high performance of our method to employ for real-world perception. On the other hand, for thermal-insensitive categories, such as traffic sign, building, bump, due to the effective visual quality preservation and enhancement, our method achieves significant superiority.  It is worth pointing out that our method can exploit and strengthen the information from different modality images. 

\subsection{Ablation studies}
\begin{figure}[!htb]
	\centering
	\setlength{\tabcolsep}{1pt} 
	
	\includegraphics[width=0.46\textwidth,height=0.11\textheight]{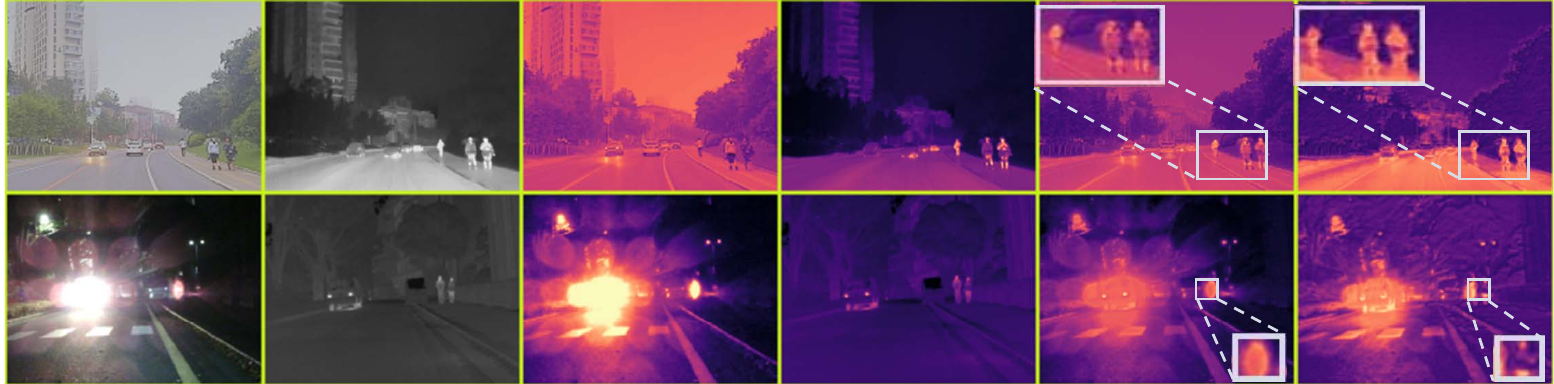}
	
	\caption{Feature visualization of different stages. From left to right: visible image, infrared image, their features, w/o HIA, and w/ HIA.}
	\label{fig:abstructure}
	\vspace{-0.2cm}  
\end{figure}

\noindent\textbf{Study on HIA.} HIA plays a key role in preserving intrinsic modality features from the semantic feature guidance. Firstly, we visualized the representative feature to discuss the effectiveness of HIA in Figure~\ref{fig:abstructure}. Clearly, HIA can remarkably preserve the salient infrared features with abundant semantic information, avoiding the interference of harsh weather and strong light. Then we plot different variants of HIA to illustrate the inner mechanisms of HIA. As shown in Figure~\ref{fig:trainings}. Obviously, the version w/o SoAM loses the classify ability to distinguish 
confused objectives, e.g., the orange circle in the second row. Meanwhile, ``w/o MoAM'' cannot protect the details at nighttime with color distortion, e.g, the building in the distance. It is worth pointing out that our full model not only provides clear visual observation but also has high sensitiveness to segmentation. Similarly, the quantitative results reported in Table~\ref{tab:strurecture1} also demonstrate the effectiveness of full HIA for both segmentation benchmarks compared with direct aggregation and other model variants. In brief, HIA is capable enough to bridge the  fusion and  segmentation tasks.

%
%

\begin{table}[!htb]
	\centering
	\renewcommand\arraystretch{1.1} 
	\setlength{\tabcolsep}{1.5mm}
	\begin{tabular}{|c|cc|cc|cc|}
		\hline
		\multirow{2}{*}{\footnotesize Model}&\multicolumn{2}{c|}{\footnotesize HIA}&\multicolumn{2}{c|}{\footnotesize MF Dataset}&\multicolumn{2}{c|}{\footnotesize FMB Dataset}\\
		\cline{2-7} 
		~&{\footnotesize SoAM}&{\footnotesize MoAM}&\footnotesize mAcc &\footnotesize mIoU &\footnotesize mAcc&\footnotesize mIoU\\
		\hline
		\footnotesize ``Concatenate''& \ding{55}  & \ding{55}&\footnotesize 72.6 &\footnotesize 52.7&\footnotesize 72.3 &\footnotesize 51.3\\\hline
		\footnotesize ``Summation''& \ding{55}  & \ding{55}&\footnotesize 71.5 &\footnotesize 52.6&\footnotesize \cellcolor{blue!15}73.7 &\footnotesize 52.1\\\hline
		\footnotesize ``Average''& \ding{55}  & \ding{55}&\footnotesize 72.9 &\footnotesize 52.1&\footnotesize 72.4&\footnotesize 50.7\\\hline
		\footnotesize M1& \ding{55}  & \ding{55}&\footnotesize67.2 &\footnotesize 51.9&\footnotesize 72.4 &\footnotesize 50.5\\
		\hline 
		\footnotesize M2& \ding{55}  & \ding{52} &\footnotesize  \cellcolor{blue!15} 73.6 &\footnotesize\cellcolor{blue!15} 55.0 &\footnotesize 72.7 &\footnotesize 52.3\\
		\hline 
		\footnotesize M3& \ding{52}  & \ding{55}&\footnotesize 72.0&\footnotesize 53.4 &\footnotesize 73.1& \footnotesize \cellcolor{blue!15}54.1 \\
		\hline 
		
		\footnotesize M4& \ding{52}  & \ding{52} &\footnotesize\cellcolor{red!15}74.8&\footnotesize \cellcolor{red!15}56.1 &\footnotesize \cellcolor{red!15}74.5&\footnotesize \cellcolor{red!15}54.8\\
		\hline   
	\end{tabular}
	
	\caption{Numerical results about the effectiveness of HIA. The first three are the results of  direct feature aggregation. Latter are the results of  model variants with HIA.}
	\label{tab:strurecture1}
\end{table}

\begin{figure}[!htb]
	\centering
	\setlength{\tabcolsep}{1pt}
	\begin{tabular}{ccccccc}
		\includegraphics[width=0.065\textwidth,height=0.115\textheight]{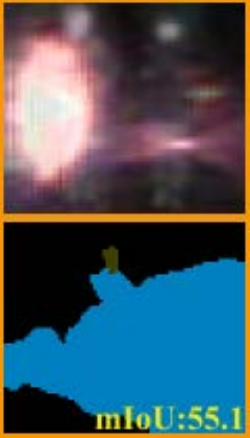}
		&\includegraphics[width=0.065\textwidth,height=0.115\textheight]{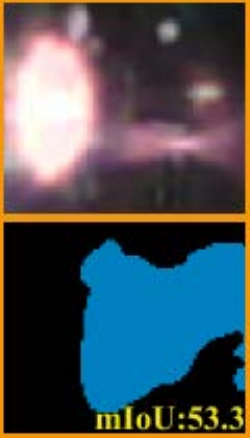}
		&\includegraphics[width=0.065\textwidth,height=0.115\textheight]{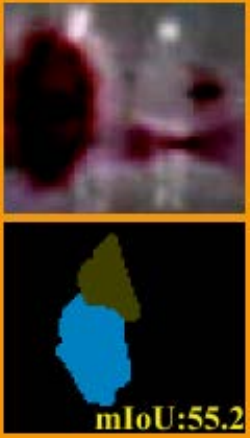}
		&\includegraphics[width=0.065\textwidth,height=0.115\textheight]{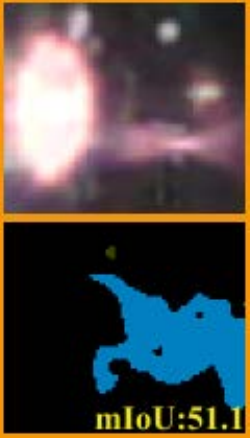}
		
		&\includegraphics[width=0.065\textwidth,height=0.115\textheight]{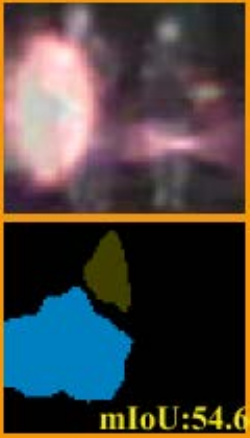}
		&\includegraphics[width=0.065\textwidth,height=0.115\textheight]{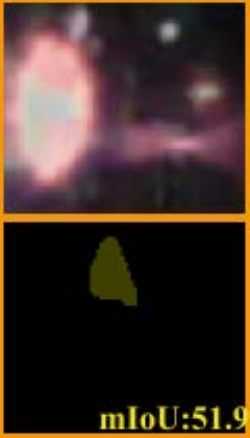}
		&\includegraphics[width=0.065\textwidth,height=0.115\textheight]{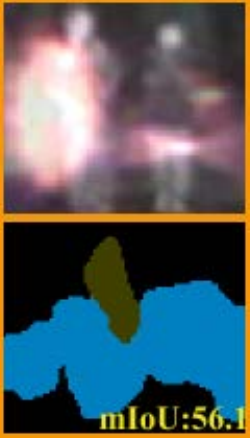}
		\\
		\footnotesize Manual&	\footnotesize DWA\cite{liu2019end}&\footnotesize GLS\cite{chennupati2019multinet++}&\footnotesize GDN\cite{chen2018gradnorm}&\footnotesize RLM\cite{lin2021closer}&\footnotesize UW\cite{kendall2018multi}&\footnotesize Ours				
	\end{tabular}
	\vspace{-0.2cm}
	\caption{Comparisons of different strategies of adjusting dynamic factors. }
	\label{fig:lamda}
\end{figure}

\noindent\textbf{Analyzing the dynamic factor.} We discussed the impact of the proposed dynamic  factor for interactive learning compared with  existing multi-task optimization methods, as shown in Figure~\ref{fig:lamda}. Manual adjustment requires plenty of prior knowledge and labor consumption. But concurrently, it achieves decent segmentation results. The other five training strategies hardly coordinate the relationship between the two tasks and fail to achieve good visualization and segmentation performance. The dynamic factor enables to better introduce task preferences into  optimization, thus achieving excellent results on both tasks.

\begin{figure}[!htb]
	\centering
	\setlength{\tabcolsep}{1pt}
	\begin{tabular}{ccccc}
		
		\includegraphics[width=0.12\textwidth,height=0.07\textheight]{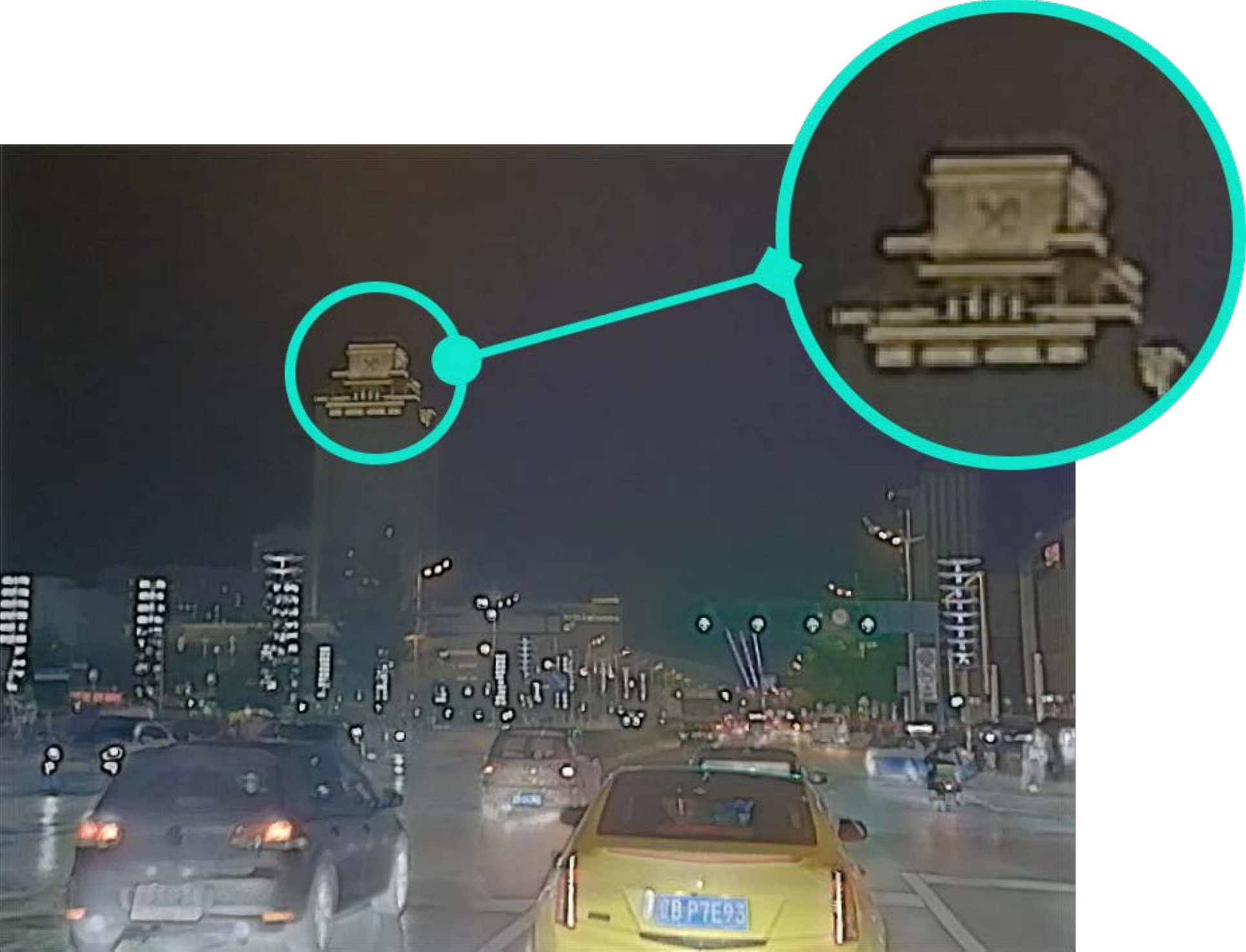}
		&\includegraphics[width=0.12\textwidth,height=0.07\textheight]{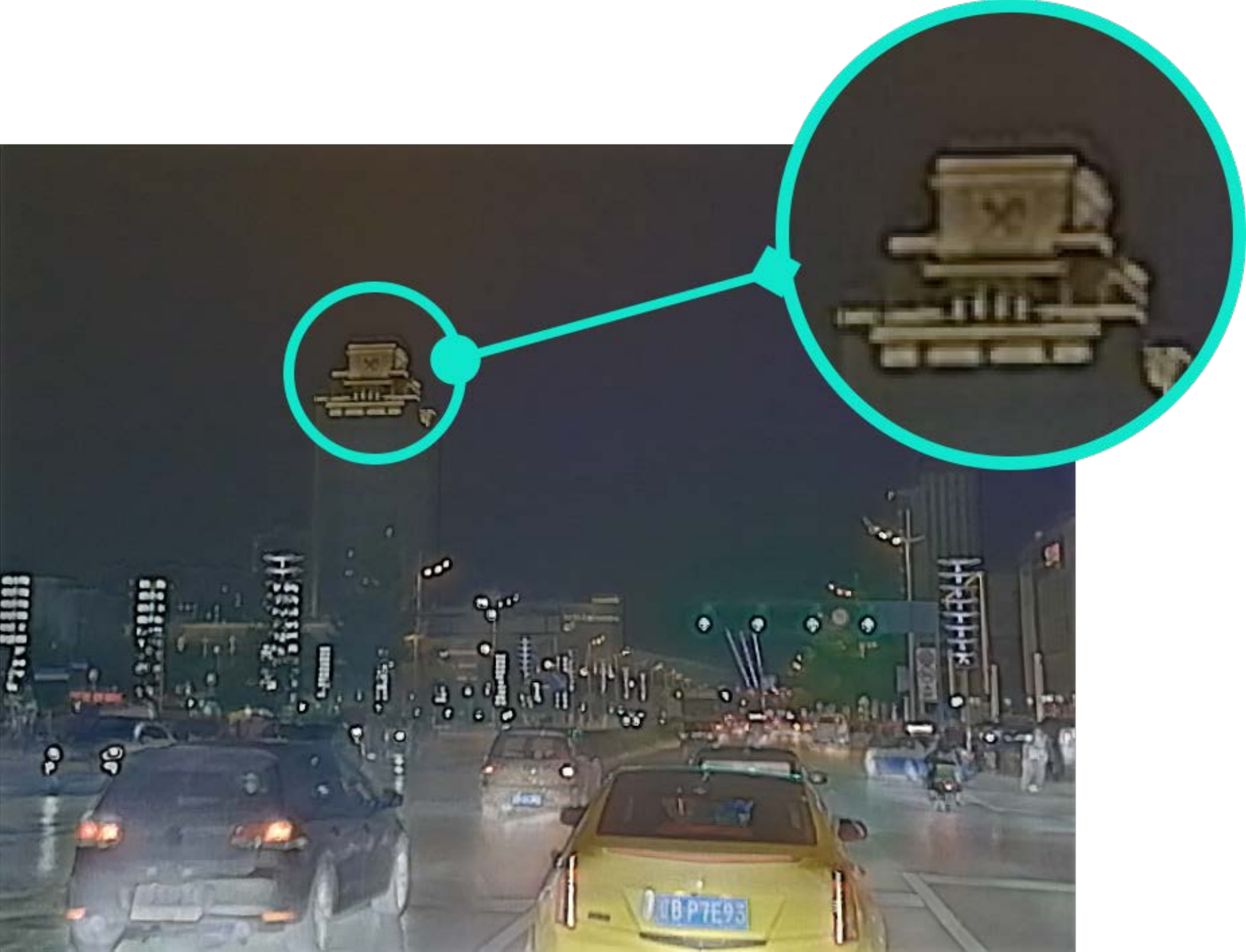}
		&\includegraphics[width=0.12\textwidth,height=0.07\textheight]{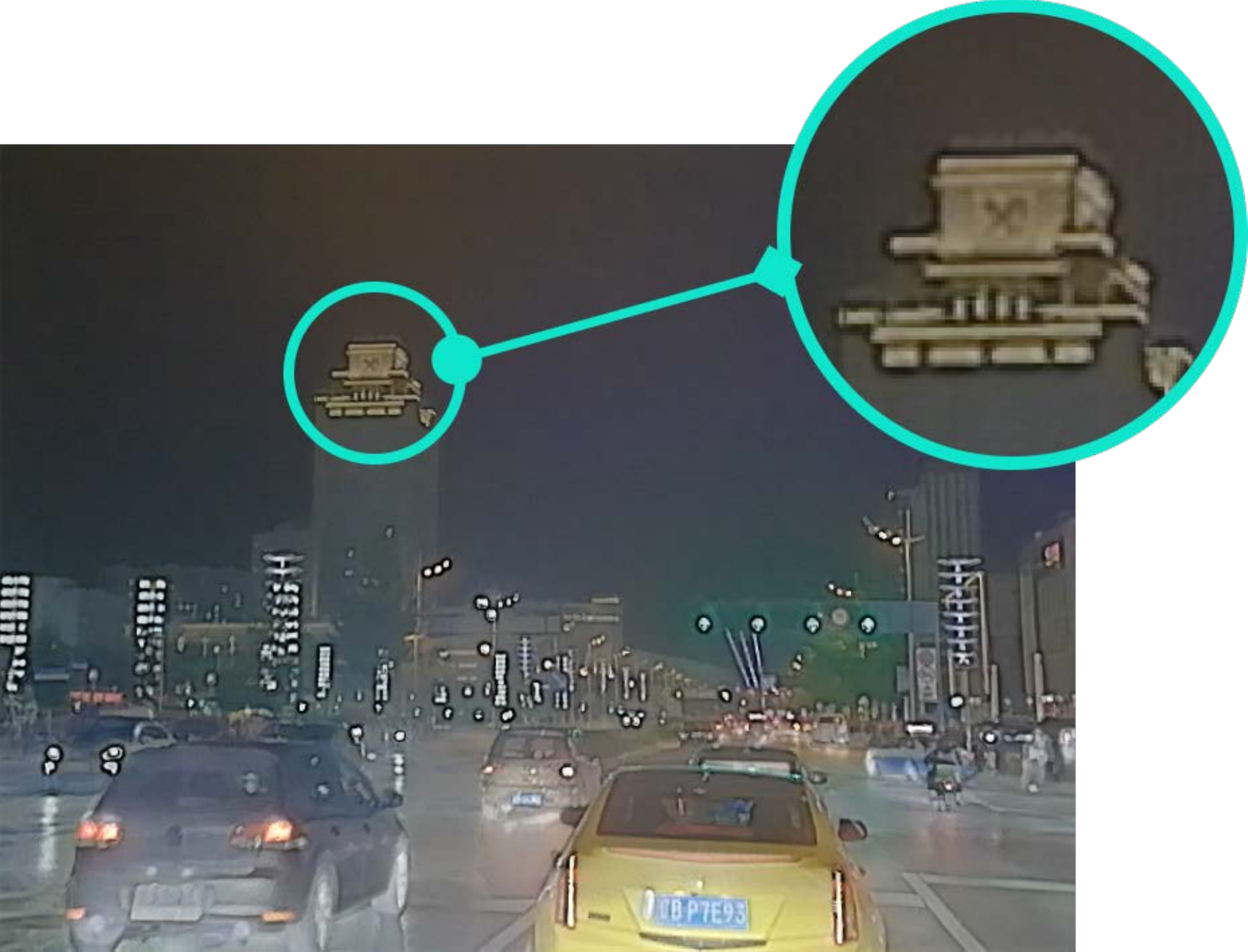}
		&\includegraphics[width=0.12\textwidth,height=0.07\textheight]{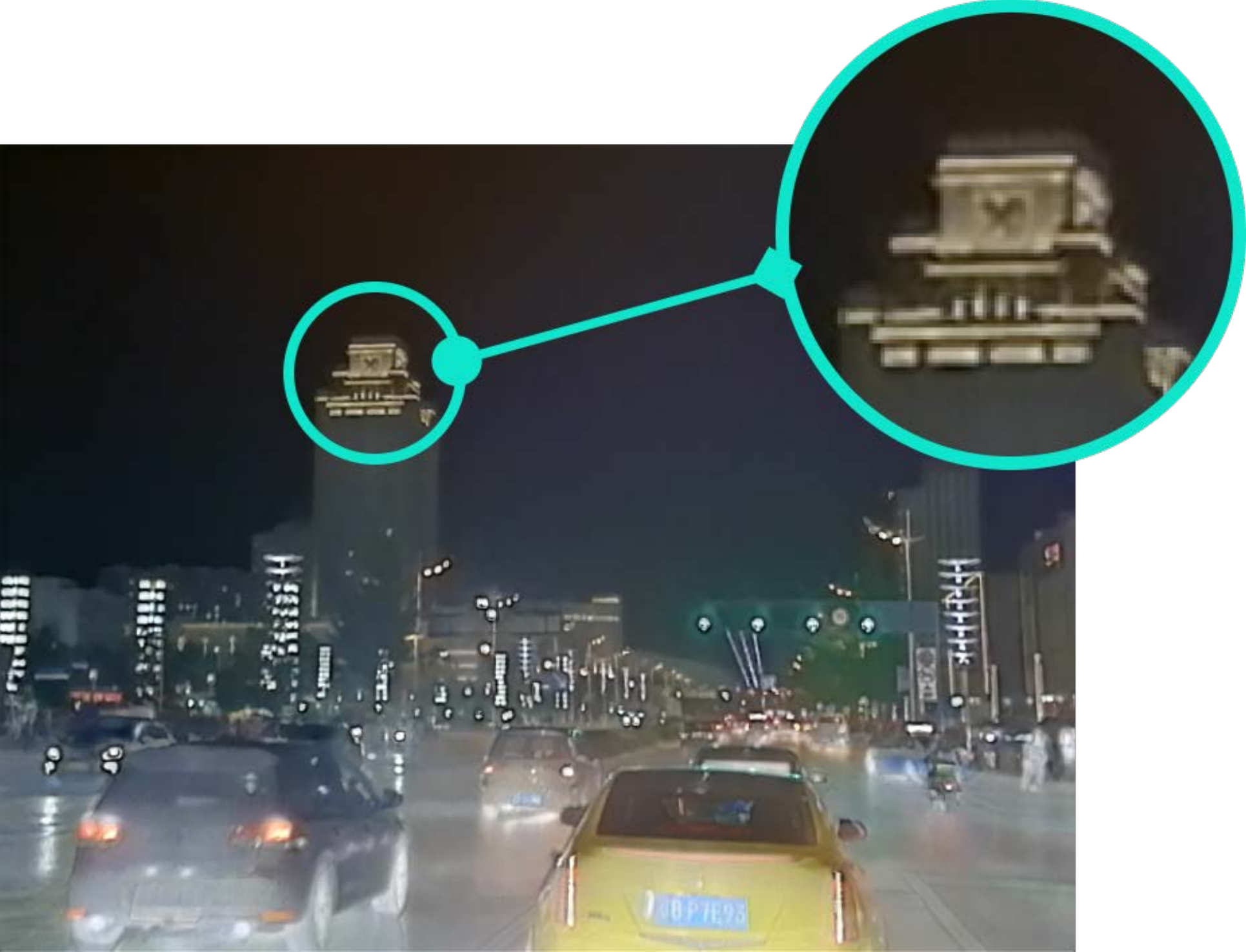}
		\\
		\includegraphics[width=0.12\textwidth,height=0.07\textheight]{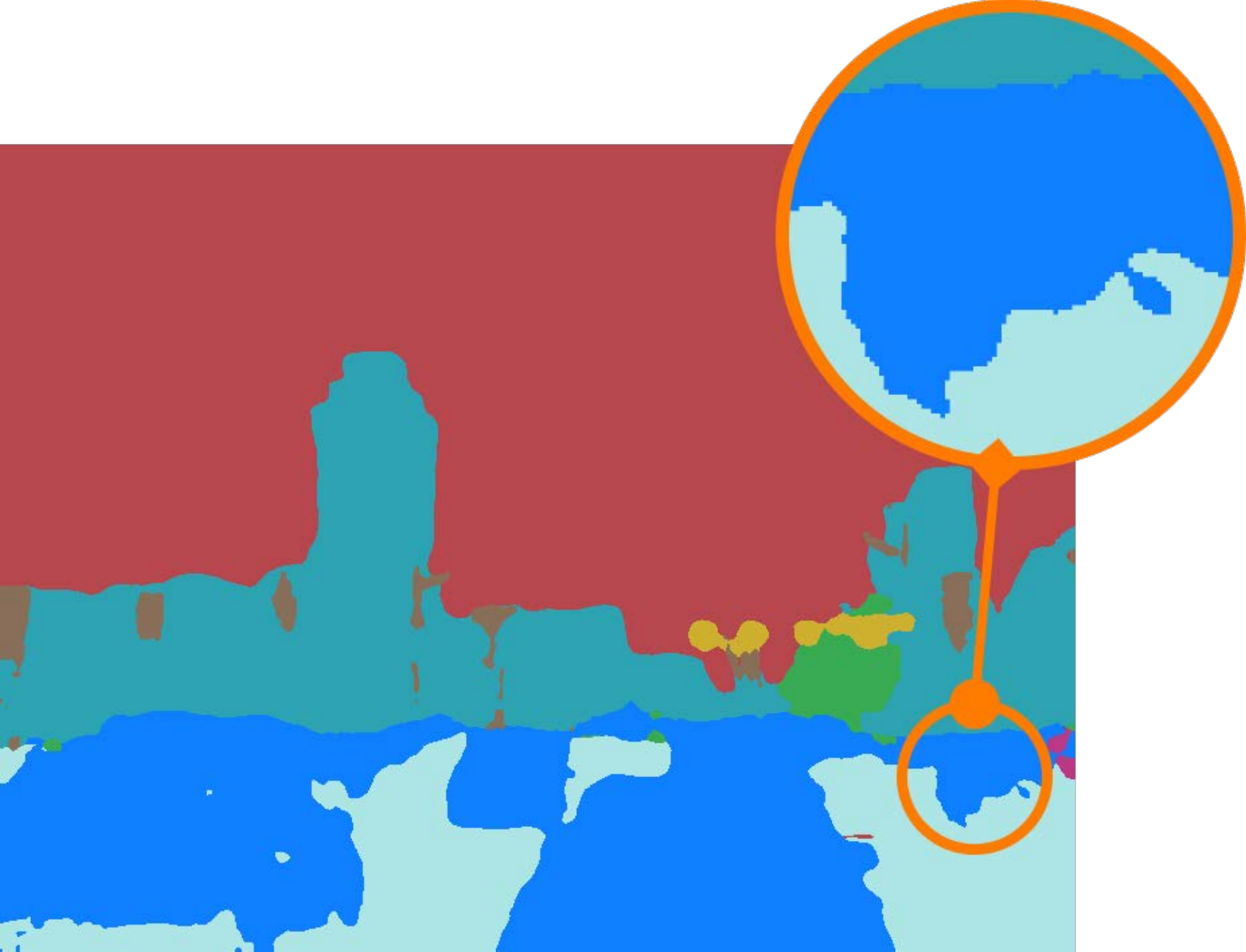}
		&\includegraphics[width=0.12\textwidth,height=0.07\textheight]{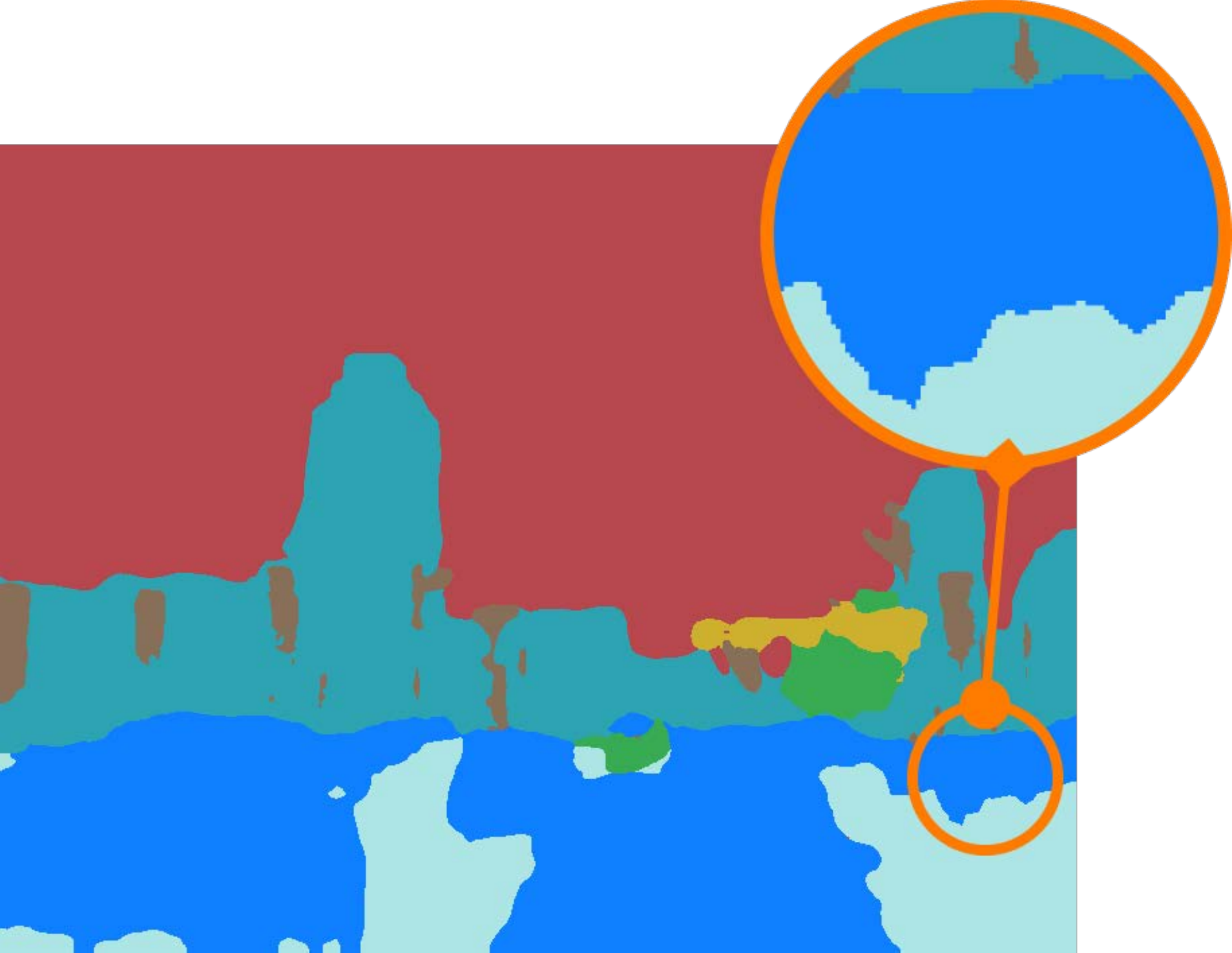}
		&\includegraphics[width=0.12\textwidth,height=0.07\textheight]{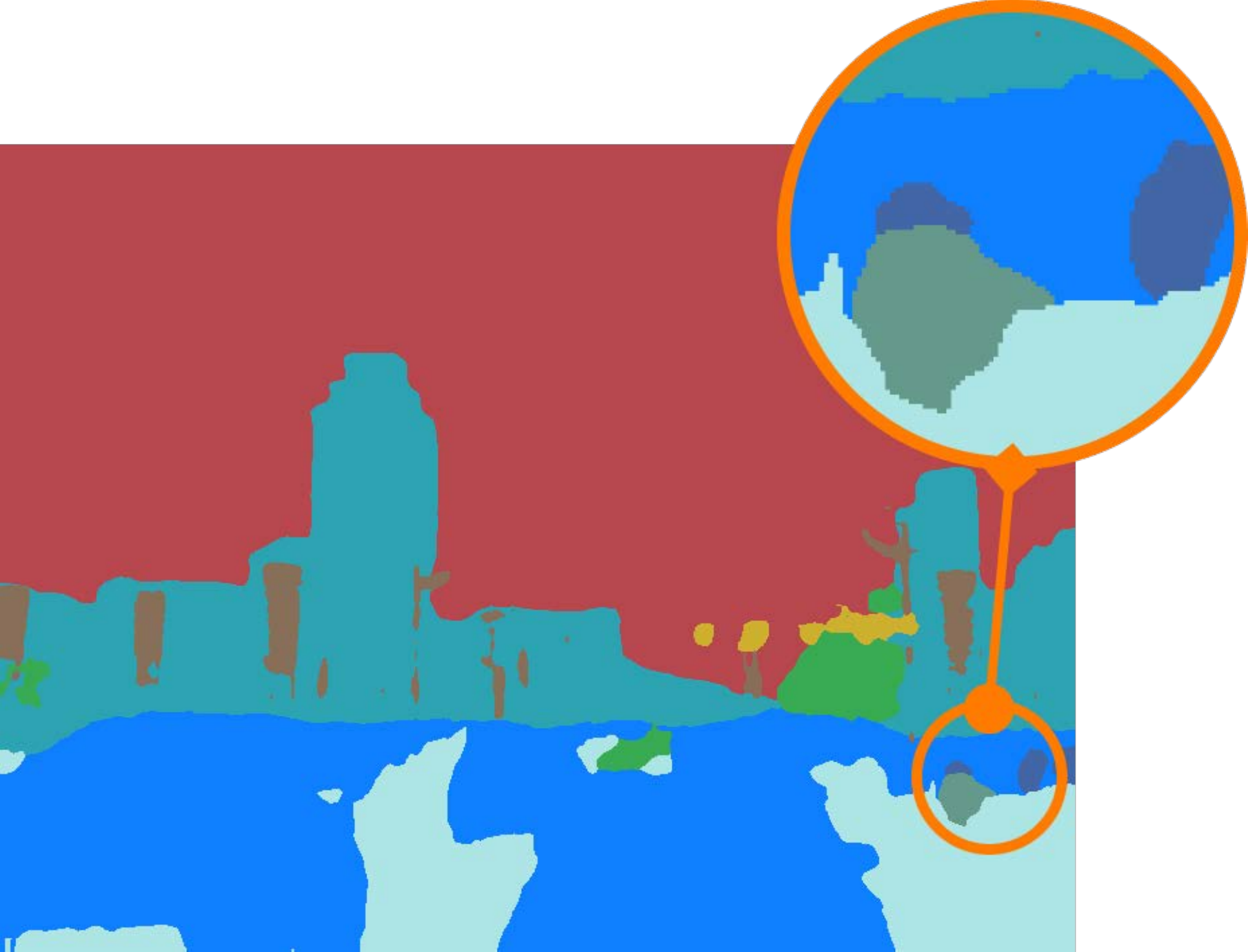}
		&\includegraphics[width=0.12\textwidth,height=0.07\textheight]{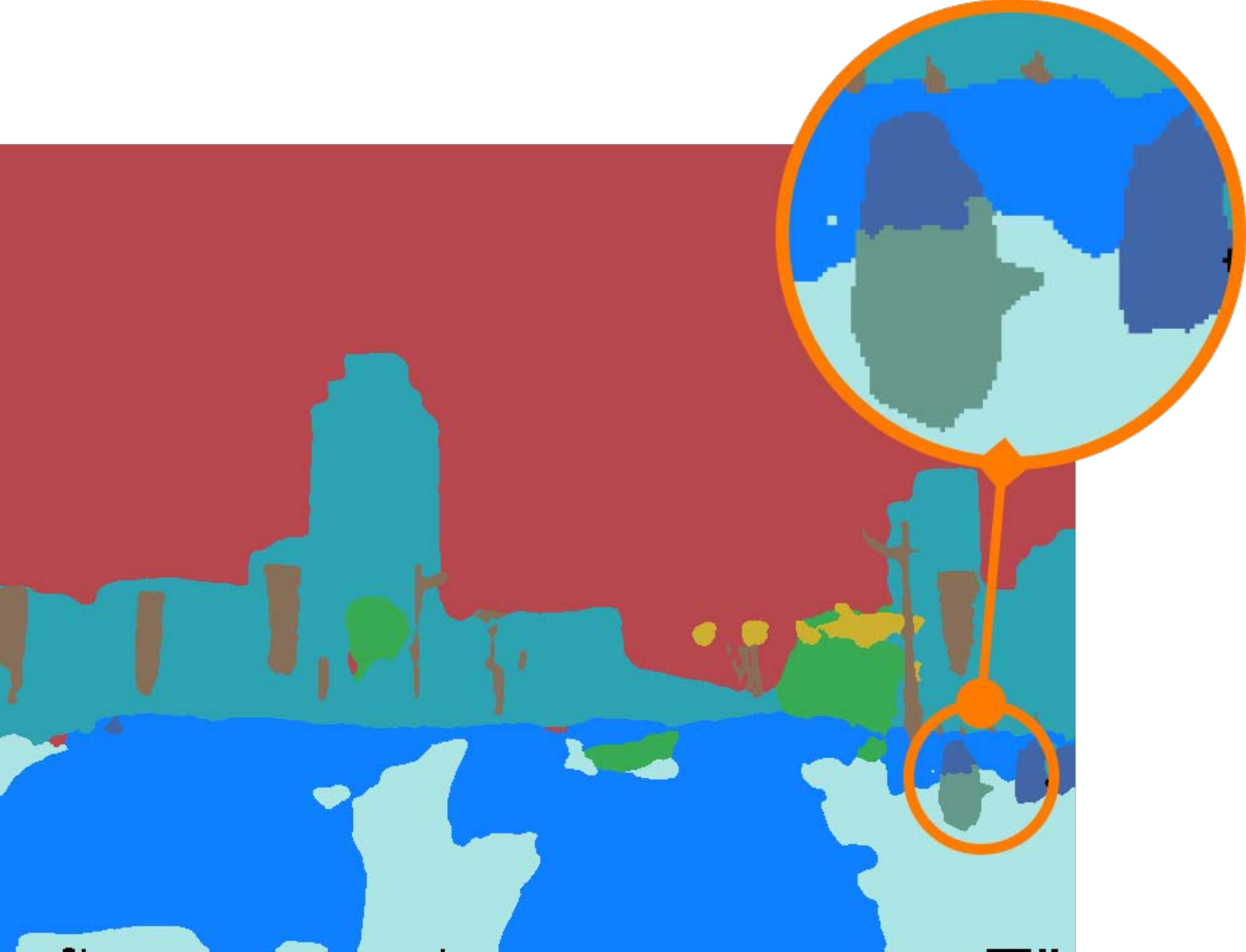}
		\\
		\footnotesize w/o HIA&\footnotesize w/o SoAM&\footnotesize w/o MoAM&\footnotesize Full Model				
	\end{tabular}
	\vspace{-0.3cm}
	\caption{Visual comparisons of different models. }
	\label{fig:trainings}
\end{figure}
\vspace{-0.2cm}

\section{Conclusion}
In this paper, a multi-interactive architecture was proposed to formulate fusion and segmentation in a harmonious manner. We introduced a hierarchical interactive attention with dynamic  factors, which bridges gaps of cross-task features from architecture and learning perspectives. In addition, we proposed a comprehensive full-time multi-modality benchmark, with well-registered targets, abundant scenes and affluent labels.

\noindent\textbf{Acknowledgments.} This work is partially supported by the National Key R\&D Program of China (No. 2022YFA1004101), the National Natural Science Foundation of China (No. U22B2052,  62027826).
{\small
	\bibliographystyle{unsrt}
	\bibliography{egbib}
}
\end{document}